\documentclass{article}
\usepackage{iclr2027_conference,times}

\usepackage{amsmath,amsfonts,bm}

\newcommand{\newterm}[1]{{\bf #1}}

\def\eqref#1{equation~\ref{#1}}

\def\1{\bm{1}}

\DeclareMathAlphabet{\mathsfit}{\encodingdefault}{\sfdefault}{m}{sl}
\SetMathAlphabet{\mathsfit}{bold}{\encodingdefault}{\sfdefault}{bx}{n}

\usepackage{hyperref}
\usepackage{url}
\usepackage{booktabs}
\usepackage{array}
\usepackage{multirow}
\usepackage{tikz}
\usetikzlibrary{positioning}
\usepackage{colortbl}
\definecolor{oursC}{gray}{0.92}
\usepackage{algorithm}
\usepackage{algpseudocode}
\usepackage{subcaption}
\usepackage{wrapfig}
\usepackage{amsthm}
\newtheorem{theorem}{Theorem}
\newtheorem{assumption}{Assumption}
\title{StateTape: Action-Conditioned Evidence Lifecycle Modeling for Long-Horizon Coding Agents}

\author{Ziyang Yu$^{1,2}$, Liang Zhao$^{1,2}$\thanks{Corresponding author.}, Bowen Zhu$^{1}$, Hasibul Haque$^{1}$ \\
$^{1}$Causal Dynamics Lab \\
$^{2}$Emory University
}

\iclrfinalcopy
\begin{document}

\maketitle

\begin{abstract}
Despite the recent success of coding agents built on large language models, it remains challenging
to run them over long horizons, since every observation is appended to the context and the context grows with each one.
\emph{History-based maintenance} is a common remedy, which masks or summarizes old observations,
or prunes what a model reads as useless, and bounds the context at little cost.
However, it decides from the text of the history alone and sees nothing of how the code is connected.
Since a coding agent edits code many times over a single task, and each write can change what
code elsewhere means, such maintenance may keep records a write has falsified, drop ones that
still hold, and miss code the agent needs next. To overcome these challenges, this paper proposes
StateTape, a novel and scalable framework that rewrites a coding agent's context as the repository
changes rather than as the context grows.
The key idea of StateTape is to model the repository as a symbol-level code graph, whose
dependencies and language rules expose which symbols a write can affect. Upon this graph, a tape
marks the symbols each write changed, which turns staleness from an inference about text into
an observation of the agent's writes. We propose a per-write procedure in which the
tape nominates the records a write could have falsified while a small manager model settles what the
write log cannot, and further provide a theoretical analysis and TraceBench, a benchmark that labels
what an agent is holding against what is actually needed. Empirically, we demonstrate that StateTape can effectively clear
falsified records and retrieve what is needed, and thus achieve a higher resolve rate in all experiments spanned by six coding agents and three edit-heavy benchmarks with little computational
overhead.
\end{abstract}

\vspace{-3mm}
\section{Introduction}
\label{sec:intro}
\vspace{-2mm}
Coding agents built on large language models have shown promising results on real software
engineering work, running for hundreds of steps on a single task where they read files, run tests
and edit code \citep{yang2024sweagent,wang2025openhands,wang2024codeact,zhang2024autocoderover}.
Despite their great promise, these agents meet significant challenges on long-horizon coding tasks that require many
edits. On benchmarks built from such tasks
\citep{swebenchpro2025,sweevo2025,swemarathon2025,slopcodebench2025}, the resolve rate declines as
the task grows longer, even when every individual step lies within the model's ability
\citep{kwa2025measuring,sinha2026illusion}. Part of this difficulty is mechanical: every
observation is appended to the context and a long
context costs tokens and degrades the model's use of what it already holds
\citep{liu2024lost}. To tackle
this challenge, context management, which decides what an agent carries from one request to the next,
has attracted fast-increasing attention and has become standard practice in agent harnesses
\citep{lindenbauer2025complexity,wang2026openhands,liu2026dive}.

\vspace{-1mm}
A key challenge in context management lies in retaining the observations that later steps depend
on while keeping the context small enough to stay affordable and usable. On the one hand, carrying
the entire history preserves every observation the agent has made, whereas the agent pays for that
history on every later request and the cost grows without bound as the task proceeds. On the other
hand, trimming the history bounds the context, while it can discard evidence the agent still needs
and force the agent to recover that evidence through additional steps. More recently, deciding what to
keep directly from the recorded history has been widely adopted and achieved strong results in
long-horizon coding agents \citep{lindenbauer2025complexity,wang2026openhands,liu2026dive,
xiao2026reducing,yi2026learning,li2026escaping}. Specifically, by masking or collapsing
observations older than a window \citep{lindenbauer2025complexity}, summarizing the history once it
passes a size limit \citep{wang2026openhands,liu2026dive}, or prompting a model to read each record
and discard the ones it judges useless or expired
\citep{xiao2026reducing,yi2026learning,li2026escaping}, such methods, which we call
\emph{history-based maintenance}, bound the context at a cost small next to the agent's own.

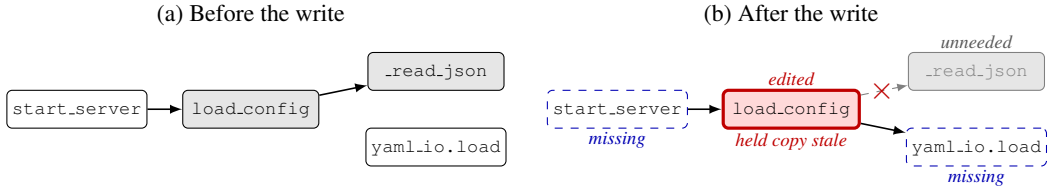
\begin{figure}[t]
\centering
\begin{tikzpicture}[
  sym/.style={draw, rounded corners=2pt, minimum height=5mm, minimum width=18mm, inner sep=2pt,
              font=\scriptsize\ttfamily},
  held/.style={sym, fill=gray!20},
  call/.style={-latex, semithick},
  tag/.style={font=\scriptsize\itshape, inner sep=1pt}]
\node[sym]  (a1) at (0,0)      {start\_server};
\node[held] (a2) at (2.3,0)    {load\_config};
\node[held] (a3) at (4.75,0.5) {\_read\_json};
\node[sym]  (a4) at (4.75,-0.5){yaml\_io.load};
\draw[call] (a1) -- (a2);
\draw[call] (a2) -- (a3);
\node[font=\small] at (2.3,1.25) {(a) Before the write};
\begin{scope}[xshift=7.15cm]
\node[sym, draw=blue!70!black, dashed]   (b1) at (0,0)      {start\_server};
\node[sym, draw=red!75!black, fill=red!15, very thick] (b2) at (2.3,0) {load\_config};
\node[sym, draw=gray, fill=gray!20, text=gray] (b3) at (4.75,0.5) {\_read\_json};
\node[sym, draw=blue!70!black, dashed]   (b4) at (4.75,-0.5){yaml\_io.load};
\draw[call] (b1) -- (b2);
\draw[-latex, gray, dashed] (b2) -- node[red!75!black, font=\large, pos=0.45] {$\times$} (b3);
\draw[call] (b2) -- (b4);
\node[tag, red!75!black, above=0pt of b2] {edited};
\node[tag, red!75!black, below=0pt of b2] {held copy stale};
\node[tag, gray!70!black, above=0pt of b3] {unneeded};
\node[tag, blue!70!black, below=0pt of b1] {missing};
\node[tag, blue!70!black, below=0pt of b4] {missing};
\node[font=\small] at (2.3,1.25) {(b) After the write};
\end{scope}
\end{tikzpicture}
\vspace{-2mm}
\caption{A single write leaves the context wrong in three ways. Boxes are functions and arrows are
calls; gray fill marks what the agent has read. (a)~The agent reads \texttt{load\_config} and its
helper \texttt{\_read\_json}. (b)~It rewrites \texttt{load\_config} to parse YAML and raise
\texttt{ConfigError}. Its held copy of \texttt{load\_config} is now \emph{stale};
\texttt{\_read\_json} has lost its only caller and is \emph{unneeded}; the caller
\texttt{start\_server} and the new callee \texttt{yaml\_io.load} are needed next but
\emph{missing} (dashed).}
\label{fig:example}
\vspace{-3mm}
\end{figure}

\vspace{-1mm}
\begin{wrapfigure}{r}{0.55\linewidth}
\vspace{-1.3\intextsep}
\centering
\includegraphics[width=\linewidth]{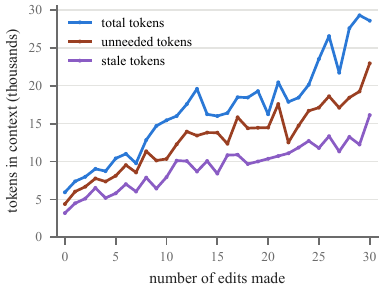}
\vspace{-1.5\intextsep}
\caption{Mean held tokens at the TraceBench decision points against the number of edits made
before them. \textbf{73.4\%} of the tokens Claude Code holds are unneeded, and \textbf{58.5\%} of
the tokens it needs are missing.}
\label{fig:stalegrowth}
\vspace{-0.6\baselineskip}
\end{wrapfigure}
However, history-based maintenance is limited by the signal it decides on: the text of the history.
What makes a coding agent different from an agent that only remembers is that its own writes change
what its evidence means. Figure~\ref{fig:example} gives an example. Consider an agent that is asked to make \texttt{load\_config} validate its
input. It reads \texttt{load\_config}, which parses the file through a helper \texttt{\_read\_json},
and it reads that helper too. It then rewrites \texttt{load\_config} so that the function parses YAML
with \texttt{yaml\_io.load} and raises \texttt{ConfigError} when a key is missing. This single write
leaves the context wrong in three ways. First, the held copy of \texttt{load\_config} is
\emph{stale}, because it shows a version that never raises and that the file no longer contains.
Second, the held copy of \texttt{\_read\_json} is \emph{unneeded}. Its text is still correct, but the
write deleted its only call, so no later step will use it. Third, the evidence for the next edit is
\emph{missing}. The function \texttt{start\_server} calls \texttt{load\_config} and must now catch
\texttt{ConfigError}, and the agent has read neither that caller nor \texttt{yaml\_io.load}.
All three errors follow from the write and the call edges around it. A mask, a summary or a pruning
model sees only records, so it cannot tell that \texttt{\_read\_json} lost its caller, and no record
names \texttt{start\_server}.
Figure~\ref{fig:stalegrowth} measures these errors at the 403 moments where real Claude Code runs are
about to edit. Unneeded tokens grow with every write rather than settling.
Every later request carries them, costing tokens and lowering the resolve rate (\S\ref{sec:experiments}).

\vspace{-1mm}
Closing this gap means deciding, at every write, what to drop and what to add, which is hard for
two reasons: 1). \emph{Deciding staleness from writes rather than from text.} A write and a held
record have to be comparable at the same granularity. Rewriting \texttt{load\_config} should mark its
held copy as stale, but it should not mark every definition in the same file, and no stale record
should be missed. 2).
\emph{Adding the evidence the next edit needs but the agent has not read.} What is missing is rarely
the edit site itself but a definition it calls, a caller that must change with it, or a contract it
must meet, all of which lie along dependency edges that a keyword search cannot enumerate. In the
example, the task never mentions \texttt{start\_server}, so a search over it misses this caller.

\vspace{-1mm}
\textbf{Our Contribution.} To jointly address all challenges above, we propose \textbf{StateTape}, a
context maintenance framework towards the trade-off between the cost and quality of the context. In StateTape, we design a symbol-level
\emph{code graph} of definitions and call edges that is updated at every write, so that writes, held
records and the dependencies among them are comparable in the same granularity. Based on this graph, we use \emph{tape} to formulate context maintenance as a
\emph{write-conditioned} procedure, where what leaves the context and what enters it are both decided
by the symbols the current write changed. 
Building on the tape, we further propose a
lightweight \emph{manager} that removes what is stale and useless in the context, and retrieves what the task needs by walking the code graph outward
from the changed symbols.
In the example, the tape marks \texttt{load\_config} as changed and refreshes its held copy. The
manager then drops \texttt{\_read\_json}, which has lost its caller, and retrieves
\texttt{start\_server} and \texttt{yaml\_io.load}, which lie one call edge from the write.
Finally, we perform extensive evaluations over five
comparison methods on three edit-heavy benchmarks with six coding agents, where StateTape attains
the best resolve rate in every case at few computational overhead.

\vspace{-4mm}
\section{Related Work}
\label{sec:related_work}

\vspace{-3mm}
\paragraph{Context management for coding agents.}
Coding agents act on a repository through tools
\citep{yang2024sweagent,wang2025openhands,zhang2024autocoderover,wang2024codeact,xia2025agentless},
and on benchmarks that need many edits
\citep{swebenchpro2025,sweevo2025,swemarathon2025,slopcodebench2025,jin2026chainswe} resolve rate
falls with horizon even when each step is within the model's competence
\citep{kwa2025measuring,sinha2026illusion}. Their context is managed by age or by a model:
collapsing or masking old observations \citep{lindenbauer2025complexity}, summarizing past a size
threshold \citep{wang2026openhands,liu2026dive}, or pruning what a model judges useless or expired
\citep{xiao2026reducing,liu2026context,wang2026swepruner}, with AdaCoM and ContextCurator
\citep{yi2026learning,li2026escaping} training a second model to edit a frozen agent's context, the
same two-model split as our manager. CORVUS
\citep{zheng2026corvus} re-injects registered files' on-disk contents every cycle, but
unconditionally, whole-file, and for file reads alone; coherence-debt measurements
\citep{mohammadi2026working} name the missing invalidation bit, and CoAgent \citep{lyu2026coagent}
supplies it only when \emph{another} agent writes the object. None of them reads the event that falsifies a
record: the agent's own write. StateTape takes that write as its signal and resolves it to the
symbols it changed, so falsified records are named rather than inferred, and the same symbols seed
retrieval for the next edit.

\vspace{-3mm}
\paragraph{Repository structure for retrieval and localization.}
Repository-level code models bring in cross-file context through structure: iterative retrieval
\citep{zhang2023repocoder}, cross-file dependencies \citep{ding2024cocomic}, and graphs over
definitions, references and dependencies
\citep{ouyang2025repograph,liu2025codexgraph,tao2025cgm,liu2024graphcoder}, which an agent may
traverse to localize code \citep{chen2025locagent} and which benchmarks score against one gold
context per task \citep{contextbench2026,corebench2026}. On the evaluation side, Slipstream \citep{chen2026slipstream} derives what a
compaction must preserve from the agent's next few steps, the nearest existing per-step target.
These works decide what to evict and what to retrieve from text or from a graph
fixed at task start, and none of them measures what the agent holds against what it still needs.
StateTape instead keeps that graph updated at every write, so that the symbols a write changed
decide both what leaves the context and what enters it, and TraceBench measures both sides of that
gap at every decision point.

\vspace{-4mm}
\section{Method}
\label{sec:method}
\vspace{-2mm}

Given a coding agent's context and the actions it takes on a repository, StateTape maintains that
context at every write: it removes evidence the task no longer needs and adds the code the next edit
is missing. We first measure, on TraceBench, how far the context an agent holds is from the context
it needs (\S\ref{sec:tracebench}), and state that gap as two objectives under one token budget
(\S\ref{sec:formulation}). StateTape consists of three components that run inside the agent harness.
First, a symbol-level code graph gives writes and held records a shared coordinate system in which
the two can be compared (\S\ref{sec:graph}). Next, the tape logs what each write changed in that
graph and nominates every held record the change could have falsified (\S\ref{sec:tape}). 
Finally, a small manager model evicts records that are stale or no longer used, and
retrieves the unread code the next edit needs by following the graph outward from the symbols that
changed (\S\ref{sec:controller}). The key idea is that the agent's own writes are the
only events that can make its evidence stale, so
staleness is observed when it happens rather than inferred from what a record says, and the same
observation locates the missing evidence. 
Figure~\ref{fig:overview} shows the procedure.

\vspace{-2mm}
\subsection{Motivation: The Held Context Is Far from the Needed Context}
\label{sec:tracebench}

\vspace{-2mm}
At every step a coding agent acts on the context it holds, $C_t$, while what it should act on is
the gold context $G_t$, the smallest body of evidence sufficient to finish the task from that step. 
As is shown in Figure~\ref{fig:partition}, the two sets split the evidence into three regions:
held but not needed, held and needed, and needed but not held.
Context management for coding agents keeps what the agent has recently seen and trims it at the
margin, which works only if the first and last regions are small (\S\ref{sec:related_work}). No
benchmark measures them, so \S\ref{sec:construction} builds TraceBench, which labels both regions at
decision points inside long, write-dense trajectories. \S\ref{sec:gap} shows that
most of what the agent holds is not needed and most of what the next edit needs is not held, the
gap StateTape addresses.

\vspace{-2mm}
\subsubsection{TraceBench}
\label{sec:construction}

\vspace{-2mm}
Existing benchmarks cannot measure this gap. Issue-resolution benchmarks score only the terminal
patch, long-context benchmarks place evidence that nothing the model does can falsify, and
localization benchmarks label the code a task needs once per task and never label what the agent
holds but no longer needs. We therefore introduce TraceBench, which
labels the held context at decision points inside long, write-dense agent trajectories. Each
decision point is annotated with both regions of Figure~\ref{fig:partition} under one protocol, and
every quantity is measured in tokens, the unit the agent pays in.

\begin{wrapfigure}{R}{0.5\linewidth}
\vspace{-1.2\intextsep}
\centering
\resizebox{\linewidth}{!}{%
\begin{tikzpicture}[x=1cm,y=1cm,>=stealth,
  gloss/.style={font=\scriptsize},
  setname/.style={font=\small}]

\definecolor{staleC}{RGB}{186,106,66}
\definecolor{liveC}{RGB}{70,128,101}
\definecolor{needC}{RGB}{100,116,134}

\fill[staleC!10] (-1.45,0) ellipse (2.75 and 2.15);
\fill[needC!9]  (1.45,0) ellipse (2.75 and 2.15);
\begin{scope}
  \clip (-1.45,0) ellipse (2.75 and 2.15);
  \fill[staleC!26] (-4.3,-2.3) rectangle (0,0.3);
\end{scope}
\begin{scope}
  \clip (-1.45,0) ellipse (2.75 and 2.15);
  \fill[liveC!20] (1.45,0) ellipse (2.75 and 2.15);
\end{scope}
\draw[staleC!65,thick] (-1.45,0) ellipse (2.75 and 2.15);
\draw[needC!65,thick]  (1.45,0) ellipse (2.75 and 2.15);

\node[staleC!85!black]        at (-2.75,1.18) {$C_t \setminus G_t$};
\node[gloss,staleC!85!black]  at (-2.75,0.72) {held, not needed};
\draw[staleC!70,dashed] (-4.173,0.3) -- (-1.273,0.3);
\node[gloss,staleC!85!black]  at (-2.75,-0.55) {$S_t$};
\node[gloss,staleC!85!black]  at (-2.75,-0.91) {stale};

\node[liveC!85!black]         at (0,0.34) {$C_t \cap G_t$};
\node[gloss,liveC!85!black]   at (0,-0.12) {held and needed};

\node[needC!85!black]         at (2.75,0.64) {$G_t \setminus C_t$};
\node[gloss,needC!85!black]   at (2.75,0.18) {needed, not held};

\node[setname,staleC!85!black] at (-2.7,-2.62) {current context $C_t$};
\node[setname,needC!85!black]  at ( 2.7,-2.62) {gold context $G_t$};

\node[gloss,staleC!85!black] at (-2.3,1.66) {(O1) evict};
\node[gloss,needC!85!black]  at ( 2.3,1.66) {(O2) retrieve};

\end{tikzpicture}%
}
\caption{Held context $C_t$ and gold context $G_t$. The stale set $S_t$ lies strictly inside
$C_t \setminus G_t$, since a true record can still be unused. O1 drains the left region and O2
pulls from the right under one token budget.}
\label{fig:partition}
\vspace{-1.6\baselineskip}
\end{wrapfigure}
We run an unmodified agent, Claude Code with Opus 4.8, on 48 SWE-EVO instances \citep{sweevo2025}
from seven repositories, so every labeled context was actually held. SWE-EVO tasks are
version-to-version feature evolutions with many-hunk patches, so runs are long and write-dense. We cut each trajectory before every edit burst, yielding 403 decision points, each labeled with
$C_t \setminus G_t$ and $G_t \setminus C_t$ (Appendix~\ref{app:tracebench}).

\vspace{-2mm}
\subsubsection{Observations}
\label{sec:gap}

\vspace{-2mm}
\paragraph{Most of what the agent holds is not needed, and the share grows with edits.}
Across the 403 decision points, 73.4\% of held tokens lie in the left region of
Figure~\ref{fig:partition} (Table~\ref{tab:tbstats}). Figure~\ref{fig:stalegrowth} shows how this
builds up as a run accumulates edits: the context grows from 5.9k tokens before the first edit to
28.6k after the thirtieth, and the part of it the imminent edit does not need grows with it from
4.4k to 23.0k tokens, 74\% to 80\% of what is held. The unneeded mass is therefore not a fixed
overhead that a larger window absorbs.

\vspace{-3mm}
\paragraph{Stale records are a majority of the held context.}
Stale records, those a later write falsified, carry 72.3\% of the left region's tokens. This measures the strict inclusion of
Figure~\ref{fig:partition}. The stale part can be identified from the agent's writes alone, since
only a write can falsify a record.

\vspace{-3mm}
\paragraph{Most of what the edit needs is not held.}
In TraceBench, 58.5\% of gold tokens are not in the context when the edit that needs them is
made, and the context covers 41.5\% of the gold context on average. What is missing is rarely the
edit site itself but the definitions, callers, and contracts a call edge away from it. Retrieval
therefore has a starting point that a query over the task description does not give it, namely the
symbols the agent is about to change.

\vspace{-2mm}
Together, the three observations point at a single signal. The unneeded side grows with the agent's
writes and is mostly stale. The missing side lies along the
call and dependency edges of the code that those writes touch. \S\ref{sec:formulation} formulates the gap as two optimization objectives and turns this
into the requirements a context management policy must meet.

\vspace{-3mm}
\subsection{Problem Formulation}
\label{sec:formulation}

\vspace{-2mm}
\paragraph{Setting and goal.}
We study context maintenance for a coding agent over a long, write-dense run on a repository. At
each write, a maintenance policy sees what the harness sees: the agent's context, the write, and the
repository before and after it, but neither the reference patch nor any label of what the task
needs. The goal is to rewrite the context so that it carries few tokens the next edit does not need
and covers the evidence that edit does need, within the agent's token budget and at a cost small
relative to the agent's own. Performance is measured at the end of a run by whether the agent
resolves the task and how many tokens it spends, and at each decision point by the two
context objectives and action-level efficiency scores defined below.

\vspace{-2mm}
\paragraph{Setup.}
A coding agent acts on a repository whose state at step $t$ is $w_t$.
Actions split into observations (reads, searches, test runs), which leave
$w_{t+1} = w_t$, and writes $\mathcal{A}_{\mathrm{wr}}$ (edits, creations,
deletions), which make $w_{t+1}$ the state $a_t$ produces from $w_t$. The agent's
context is a finite set of records $C_t$. A record $r = (c_r, \rho_r, \tau_r)$ holds
text $c_r$, a referent $\rho_r$ naming the part of the repository that
text is about, and the step $\tau_r$ at which it entered the context, and costs
$n(r) = |c_r|$ tokens, extended to sets by $n(S) = \sum_{r \in S} n(r)$. A referent
may be a byte range, a symbol, the solution set of a query, or the repository state a
test ran under.
\vspace{-1mm}

At step $t$, let the \newterm{gold context} $G_t$ be a smallest set of records
sufficient to complete the task from $t$. Its overlap with the held context defines
coverage,
$\mathrm{Rec}(C_t) = n(C_t \cap G_t) / n(G_t) \in [0,1]$, while
$\sigma(C_t) = n(C_t \setminus G_t)$ is the \newterm{unneeded mass}. A maintenance
policy acts at each write by removing $D_t \subseteq C_t$ and adding
$A_t$, where $A_t \cap C_t = \emptyset$, producing
$C^{+}_t = (C_t \setminus D_t) \cup A_t$; the agent's next observation extends this
context into $C_{t+1}$. The main agent's token budget $B$ bounds both contexts.
\vspace{-1mm}

Each record is a timestamped claim: its text says something about the repository
that was true when the agent observed it and that a later edit can make false. That
claim is a predicate $\phi_r$ on repository states, where $\phi_r(w)$ asks whether the
record still holds of $w$. Record $r$ is \newterm{stale} at $t$ once its claim has
become false, $\neg\phi_r(w_t)$; the stale set is
$S_t = \{ r \in C_t : \neg\phi_r(w_t) \}$, containing $n(S_t)$ stale tokens.

\vspace{-1mm}
\paragraph{Staleness is action-conditioned and monotone.}
Observations leave $w_t$ fixed, so writes are the only events that can falsify a
record: if no write $a_s \in \mathcal{A}_{\mathrm{wr}}$ with $\tau_r \le s < t$
touches $\rho_r$, then $\phi_r(w_t) = \phi_r(w_{\tau_r}) = \mathrm{true}$. Two things follow. First, every stale record traces back to the
write that falsified it, and that write is an event the harness already observes, so
staleness need not be guessed from the text of the record. Second, without drops or
rebuilds the stale set only grows:
\begin{equation}
S_t \;\subseteq\; S_{t+1},
\qquad\text{so}\qquad
n(S_t) \;\le\; n(S_{t+1}).
\label{eq:monotone}
\end{equation}
The gap between what the agent believes and what the repository holds therefore never
shrinks on its own, and it grows with the number of writes. A run with few writes
builds up almost none of it. This is the precise sense in which evidence lifecycle is
a long-horizon problem rather than ordinary context management, and it is why the
effects in \S\ref{sec:experiments} follow the number of edits rather than of steps.

\vspace{-3mm}
\paragraph{Optimization objective.}
The general context-maintenance problem is to reduce held but unneeded evidence
\citep{xiao2026reducing,liu2026context} and increase coverage of needed repository
evidence \citep{zhang2023repocoder} under the agent's token budget. At each write,
we express these two aims as
\begin{equation}
\min\ \sigma(C_t)
\qquad\qquad
\max\ \mathrm{Rec}(C_t)
\qquad\qquad \text{s.t.}\quad n(C_t) \le B ,
\label{eq:surrogates}
\end{equation}
The first targets the left region of Figure~\ref{fig:partition} and the second targets the right-region
deficit through gold-context coverage. These are the two gaps
measured in \S\ref{sec:gap}, stated as objectives.

\vspace{-3mm}
\paragraph{From objectives to scoring surrogates.}
\label{sec:metrics}
Eq.~\ref{eq:surrogates} states the two aims, but a policy cannot optimize it as written.
Eq.~\ref{eq:surrogates} describes a good context
and not the action that produces one. At a write the policy chooses $D_t$ and $A_t$, and
every pair with $n(C^{+}_t) \le B$ is feasible. An eviction that clears unneeded tokens
usually destroys needed ones too, and an addition that raises coverage adds
unneeded ones.

\vspace{-1mm}
We therefore score the action, not the context it leaves, and charge every action for its cost. \newterm{Collateral efficiency} measures the unneeded tokens an
eviction $D_t$ clears per needed token it destroys,
\begin{equation}
\mathrm{CE}(D_t) \;=\; n(D_t \setminus G_t) \,/\, n(D_t \cap G_t),
\label{eq:ce}
\end{equation}
and \newterm{retrieval efficiency} measures the share of the right region an addition $A_t$
recovers per token it adds,
\begin{equation}
\mathrm{RE}(A_t) \;=\; \kappa \cdot
n\bigl(A_t \cap (G_t \setminus C_t)\bigr) \,/\, \bigl(n(G_t \setminus C_t) \cdot n(A_t)\bigr),
\label{eq:re}
\end{equation}
where $\kappa > 0$ is a scale constant. It does not change the maximizer of $\mathrm{RE}$ and
only sets the unit in which $\mathrm{RE}$ is reported. The two aims then become
objectives over the action taken at each write,
\begin{equation}
\textbf{(O1)}\quad \max\nolimits_{D_t}\ \mathrm{CE}(D_t)
\qquad\qquad
\textbf{(O2)}\quad \max\nolimits_{A_t}\ \mathrm{RE}(A_t)
\qquad\qquad \text{s.t.}\quad n(C^{+}_t) \le B .
\label{eq:action}
\end{equation}

\vspace{-3mm}
\paragraph{Technical requirements.}
Approximating O1 and O2 from observable signals at every write requires addressing
three requirements, one for each observation of \S\ref{sec:gap}.
\textbf{Requirement 1 (Write-grounded staleness):} decide which held records
are stale from the writes that falsified them, without a model call and without letting a stale
record escape, since the stale set grows with every write (Eq.~\ref{eq:monotone}). \textbf{Requirement 2 (Forward-looking eviction):} remove records that are still true but
no longer used, the smaller part of $C_t \setminus G_t$, which depends on where the task goes next. 
\textbf{Requirement 3 (Dependency-targeted retrieval):} add the
missing part of $G_t$, which lies along the call and dependency edges of the code being edited.
Existing context management does not address these requirements. 
StateTape assigns these responsibilities to its
three components: the code graph gives Requirements 1 and 3 a shared coordinate system
(\S\ref{sec:graph}), the tape meets Requirement 1 without a model (\S\ref{sec:tape}), and the manager meets
Requirements 2 and 3 on top of the tape (\S\ref{sec:controller}).

\vspace{-2mm}
\subsection{StateTape}
\label{sec:overview}
\vspace{-2mm}

\begin{wrapfigure}[26]{r}{0.55\linewidth}
\vspace{-1.3\intextsep}
\centering
\includegraphics[width=\linewidth,trim={160 56 160 160},clip]{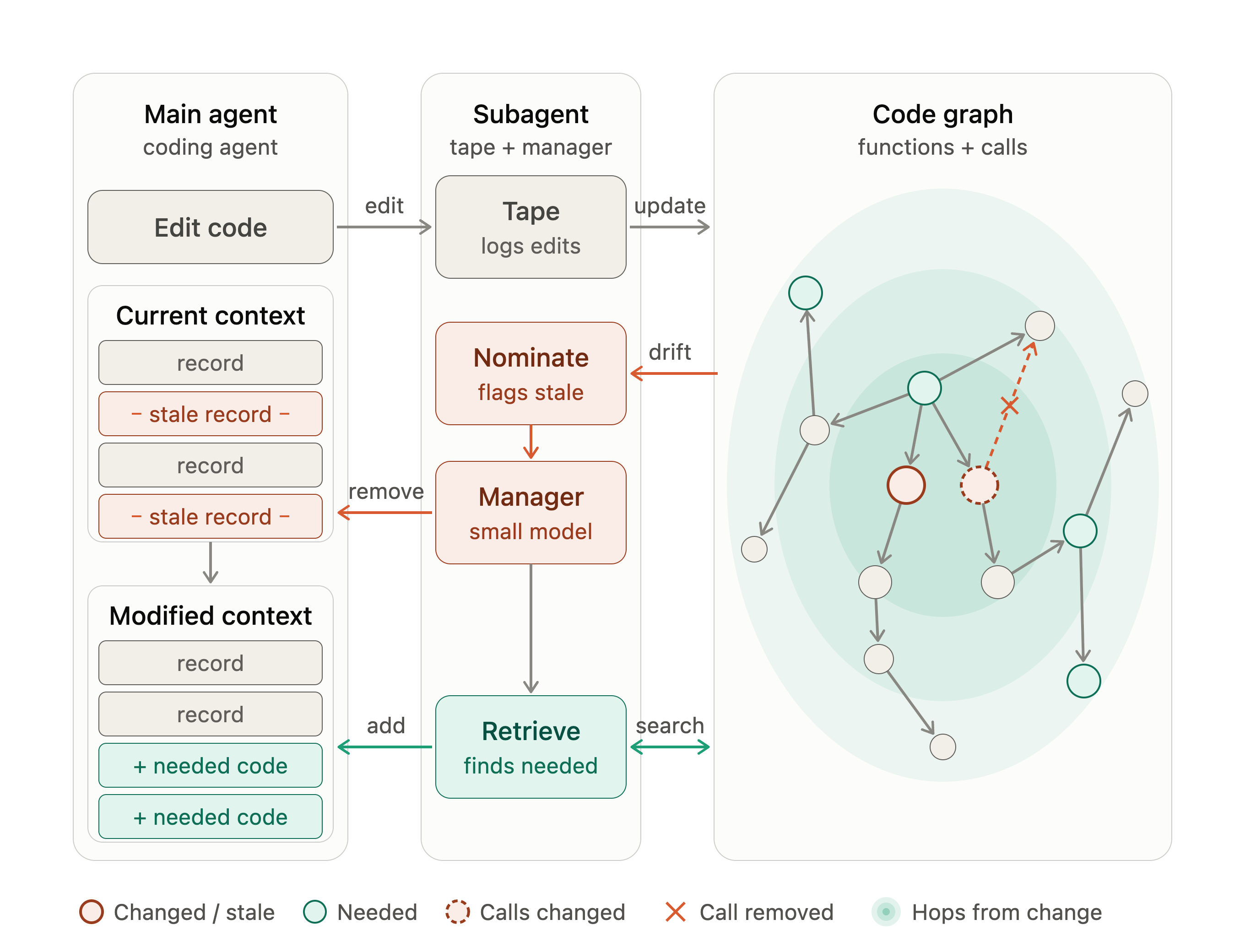}
\vspace{-1.5\intextsep}
\caption{Overview of StateTape. \emph{Left:} the agent edits code, and its context, which holds
stale records, becomes a modified context with those records removed and needed code added.
\emph{Middle:} the tape applies each edit to the code graph; the drifted symbols
nominate the records they falsify, and the manager removes records and retrieves code. \emph{Right:} the code graph of functions and calls. A write changes a function's body (solid
red) or where a call resolves (dashed; removed call crossed out); shaded rings mark hops from
the change, and green nodes are code the next edit needs. Red paths serve eviction (O1), green
paths retrieval (O2).}
\label{fig:overview}
\vspace{-2.0\baselineskip}
\end{wrapfigure}

Figure~\ref{fig:overview} provides the detailed end-to-end flow of StateTape. The tape receives the
write before it reaches the repository and applies it to the code graph; the symbols that drifted
return as nominations of the records they falsify, and the manager acts on those nominations and
searches the graph outward from the drifted symbols for the code the next edit needs. The agent's
next request then carries a maintained context rather than the one it accumulated.

\vspace{-2mm}
\subsubsection{A Symbol-Level Code Graph}
\label{sec:graph}
\vspace{-2mm}

Requirement 1 decides staleness from writes, and Requirement 3 finds missing evidence along
dependencies. Both need writes, records and dependencies in shared units, so a
write can be matched against the records it affects and the code depending on it; the code graph
provides them.

\vspace{-1mm}
The code graph describes the repository as named definitions and the calls between them. Its unit is the symbol (a function, method or class), the smallest unit whose change alters
meaning, not position.

\vspace{-1mm}
Formally, the code graph of a repository state at step $t$ is
$\Phi(t) = (V_t, E_t, b_t)$ with $E_t \subseteq V_t \times V_t$,
where $V_t$ is the set of symbols, $E_t$ holds call edges, and $b_t(v)$
is the body text of symbol $v$.

\vspace{-1mm}
A symbol keeps its identity when an edit moves it, and two states can be compared symbol by symbol through $V_t$ and
$b_t$. The edges $E_t$ carry the relationships between symbols.

\vspace{-2mm}
\subsubsection{The Tape: Action-Conditioned Staleness Checker}
\label{sec:tape}
\vspace{-2mm}

The tape decides which held records are stale (Requirement~1). At every write,
it compares the code graph before and after the write, logs the symbols that
changed, and nominates every held record that covers one of them. Because it
only compares two graphs, it makes no model call.

\vspace{-1mm}
A write can change a symbol in two ways. It can edit the symbol's body, or it
can leave the body untouched and change which definition one of the symbol's
calls reaches. Suppose the agent has read \texttt{read\_config}, which calls
\texttt{load}. If the agent then changes the import from \texttt{json\_io.load}
to \texttt{yaml\_io.load}, the body of \texttt{read\_config} stays the same, but
the record of \texttt{read\_config} no longer describes what the code does.
The tape detects the first through symbol bodies and the second through call
edges; a symbol changed either way has \newterm{drifted} (Appendix~\ref{app:examples}). 

\vspace{-1mm}
Formally, with $\Phi(t) = (V_t, E_t, b_t)$ as in \S\ref{sec:graph}, the write
$a_t$ produces two sets of drifted symbols,
\begin{equation}
\begin{aligned}
\mathcal{X}_t &\;=\; \bigl(V_{t-1} \setminus V_t\bigr) \;\cup\; \bigl\{\, v \in V_{t-1} \cap V_t \;:\; b_t(v) \neq b_{t-1}(v) \,\bigr\}, \\[2pt]
\partial\mathcal{X}_t &\;=\; \Biggl(\,\bigcup\nolimits_{(x,y) \,\in\,
\left(E_{t-1} \setminus E_t\right) \cup \left(E_t \setminus E_{t-1}\right)} \bigl\{x,y\bigr\}\,\Biggr)
\,\cap\, \bigl(V_{t-1} \cap V_t\bigr) \,\setminus\, \mathcal{X}_t .
\end{aligned}
\label{eq:driftsets}
\end{equation}
$\mathcal{X}_t$ holds the symbols the write removed or whose body it edited.
$\partial\mathcal{X}_t$ holds the remaining symbols that gained or lost a call
edge; in the example above, it contains \texttt{read\_config}. Both sets are empty
after an observation, since an observation leaves the graph unchanged. The tape
logs them with the step $t$, so each change stays attributed to the write that
made it.

\vspace{-1mm}
To check a record, the tape collects its drift since $\tau_r$,
$\mathcal{X}_{(\tau_r,t]} = \bigcup_{\tau_r < u \le t} \mathcal{X}_u$,
and $\partial\mathcal{X}_{(\tau_r,t]}$ likewise; a record is \newterm{nominated}
when it covers a symbol in either,
\begin{equation}
\mathcal{N}_t \;=\; \bigl\{\, r \in C_t \;:\;
\rho_r \cap \bigl(\mathcal{X}_{(\tau_r,\,t]} \cup \partial\mathcal{X}_{(\tau_r,\,t]}\bigr) \neq \emptyset \,\bigr\}
\;\;\supseteq\;\; S_t .
\label{eq:nominations}
\end{equation}
The inclusion $\mathcal{N}_t \supseteq S_t$ states that no stale record escapes
nomination; Theorem~\ref{thm:tape} gives the condition under which it holds.
A nominated record need not be stale; the manager decides its fate
(\S\ref{sec:controller}).

\vspace{-1mm}
The tape nominates by rule, not by a model, for two reasons. First, the rule is
exact. The code graph expresses writes and held records in the same symbols, so
each write is matched directly against the records that cover what it changed.
Second, it costs no inference (Theorem~\ref{thm:cost}).

\vspace{-2mm}
\subsubsection{The Manager}
\label{sec:controller}

\vspace{-2mm}
The manager handles what the tape cannot: records that are still true but no
longer used (Requirement~2), and code the next edit needs but the agent has not
read (Requirement~3). It is a small LLM, invoked in the agent's request path
only when the tape carries a nomination, and it sees the repository only through read-only skills
that the graph and tape answer without a model call (Table~\ref{tab:skills}).

\vspace{-1mm}
At each invocation the manager makes three decisions. First, it assigns each
nominated record one of four actions. \textsc{refresh} replaces the record with
the current body $\mathrm{fresh}(r)$. \textsc{drop} leaves a short
tombstone $\mathrm{tomb}(r)$ naming the changed symbols, so the agent knows what
it lost. \textsc{rerun} replaces a test result with a stub $\mathrm{stub}(r)$
asking the agent to run the test again. \textsc{keep} leaves a record that was
nominated but is still correct. Second, it evicts records that no write
falsified but that the task has finished with. We write this set $D^{m}_t$; the
tape cannot find it, because it depends on where the task goes next. Third, it
retrieves the code the next edit needs, $R^{m}_t$. Since that code usually lies
one call edge from the code being changed (\S\ref{sec:gap}), the manager follows
call edges outward from the drifted symbols for at most two hops and checks each
candidate against source.

\vspace{-1mm}
In the running example, the body of \texttt{read\_config} is unchanged, so
refreshing it would restore the same text. The manager drops it instead, and the
tombstone tells the agent that the call now resolves elsewhere. It also evicts
an old test output on an unrelated module, and retrieves the new callee
\texttt{yaml\_io.load} with a caller of \texttt{read\_config} that
indexes the returned dictionary.

\vspace{-1mm}
Sorting nominations by action into $N^{\textsc{drop}}_t$,
$N^{\textsc{refresh}}_t$, $N^{\textsc{rerun}}_t$ and $N^{\textsc{keep}}_t$, the
policy is
\begin{equation}
\begin{aligned}
D_t &\;=\; N^{\textsc{drop}}_t \;\cup\; N^{\textsc{refresh}}_t \;\cup\; N^{\textsc{rerun}}_t \;\cup\; D^{m}_t, \\[3pt]
A_t &\;=\; \mathrm{tomb}\bigl(N^{\textsc{drop}}_t\bigr) \;\cup\; \mathrm{fresh}\bigl(N^{\textsc{refresh}}_t\bigr) \;\cup\; \mathrm{stub}\bigl(N^{\textsc{rerun}}_t\bigr) \;\cup\; R^{m}_t .
\end{aligned}
\label{eq:update}
\end{equation}
The first line serves O1, the second O2, and the manager trades them off within $B$.

\vspace{-2mm}
\subsection{Theoretical Analysis}
\label{sec:theory}

\vspace{-2mm}
\paragraph{Soundness.}
We analyze what the tape guarantees under any manager (proofs in
Appendix~\ref{app:proofs}).

\vspace{-1mm}
\begin{theorem}
\label{thm:tape}
Under the symbol-locality condition of Assumption~\ref{asm:local}, at every write step $t$:
(a) every stale read, edit or test result in $C_t$ is in $\mathcal{N}_t$;
(b) for any manager output whose retrievals cover no drifted symbol, the stale mass of reads,
edits and test results in the maintained context $C^{+}_t$ is at most
$n(N^{\textsc{keep}}_t \cap S_t)$.
\end{theorem}

\vspace{-1mm}
Theorem~\ref{thm:tape} says that the tape finds every held record the agent's writes broke
without calling a model, and that no manager decision can reintroduce a stale record.

\vspace{-2mm}
\paragraph{Cost.}
Here we analyze the cost of StateTape over a run of $T$ requests to the main agent, $W$ of which
follow a write, with $\mathcal{T}_N$ the requests that carry a nomination.

\vspace{-1mm}
\begin{theorem}
\label{thm:cost}
(a) The tape makes no model calls, and per write it costs
$\mathcal{O}\bigl(d \sum_{r \in C_t} |\rho_r|\bigr)$, with $d$ the largest out-degree of a held symbol, beyond parsing the written
files, independent of repository size. (b) If a manager call costs
$\mathcal{O}(|\mathcal{N}_t|)$ tokens, the manager's tokens $K^{\mathrm{mgr}}$ and the main
agent's tokens $K^{\mathrm{agt}}$ satisfy
$K^{\mathrm{mgr}}/K^{\mathrm{agt}} = \mathcal{O}\bigl(|\mathcal{T}_N|\,\bar{N} / (T\, n_{\min})\bigr)$,
where $\bar{N}$ is the mean nomination count over $\mathcal{T}_N$ and $n_{\min}$ the smallest
context sent. (c) If each write falsifies at least $\delta$ held tokens, the stale tokens sent to
the agent over the run are at least $\Omega(\delta W^{2})$ without maintenance and at most
$\sum_t n(N^{\textsc{keep}}_t \cap S_t)$ with StateTape.
\end{theorem}

Theorem~\ref{thm:cost} shows that StateTape's overhead is bounded.

\vspace{-3mm}
\section{Experiments}
\label{sec:experiments}

\vspace{-2mm}
\subsection{Experimental Setup}

\vspace{-2mm}
\paragraph{Models and Datasets.}
We evaluate six main agents: Claude Haiku 4.5, Sonnet 5 and Opus 4.8 in Claude Code, and
GPT-5.6 luna, terra and sol in the Codex CLI. The manager is Haiku 4.5 for the Claude agents and
GPT-5.4 nano for the GPT agents. Resolve rate is measured on three benchmarks whose tasks require
many edits: SWE-EVO \citep{sweevo2025}, SlopCodeBench \citep{slopcodebench2025} and SWE-bench Pro
\citep{swebenchpro2025}. O1 and O2 are scored with CE and RE (\S\ref{sec:metrics}) on TraceBench (\S\ref{sec:tracebench}).
With $\kappa = 10^{5}$, RE is in percentage points of $G_t \setminus C_t$ recovered per
thousand tokens added. Besides the unmodified agent (Control), we compare four baselines in the same shim: \textbf{observation masking} and \textbf{LLM summary}
\citep{lindenbauer2025complexity}, \textbf{CORVUS} \citep{zheng2026corvus}, and a
\textbf{text-only manager} in the style of AgentDiet \citep{xiao2026reducing}
(Appendix~\ref{app:impl}).

\vspace{-3mm}
\subsection{Main Results}

\vspace{-2mm}
\paragraph{StateTape improves resolve rate for every main agent.}
StateTape has the highest resolve rate in all 18 agent--benchmark cells of Table~\ref{tab:main}.
Averaged over agents, it adds 4.9 points over Control on SWE-EVO, 2.7 on SlopCodeBench
and 3.3 on SWE-bench Pro. Sonnet 5 gains most (6.2, 3.6, 4.9).

\vspace{-3mm}
\paragraph{StateTape has little computational overhead.}
Across the 18 cells of Table~\ref{tab:cost}, the manager uses 0.8\% to 4.7\% as many tokens as the
main agent, and 2.6\% when summed over all of them.

\begin{figure}[t]
\centering
\begin{subfigure}[t]{0.49\linewidth}
\centering
\includegraphics[width=\linewidth]{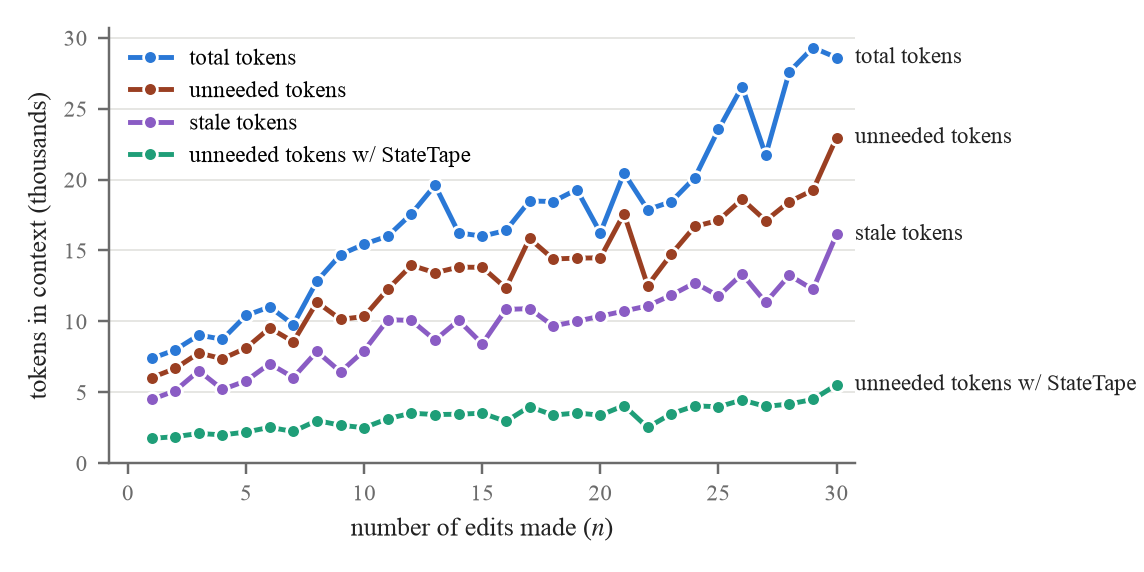}
\label{fig:staleremoved}
\end{subfigure}\hfill
\begin{subfigure}[t]{0.49\linewidth}
\centering
\includegraphics[width=\linewidth]{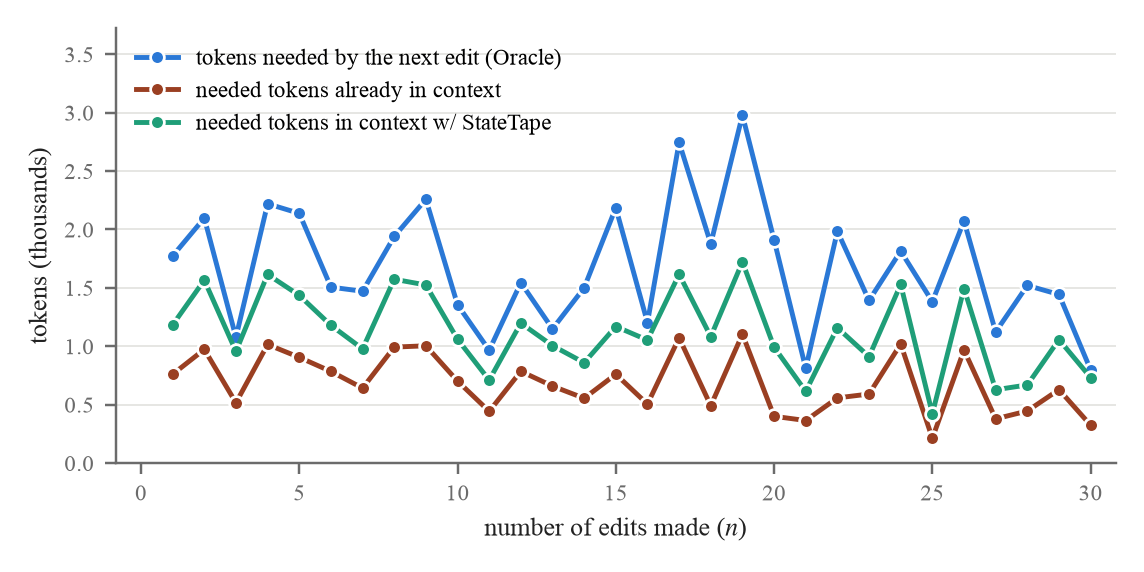}
\label{fig:neededretrieved}
\end{subfigure}
\vspace{-5mm}
\caption{Context eviction and retrieval on TraceBench as a function of the number of edits made
at each decision point. \emph{Left:} total held
tokens $n(C_t)$, unneeded tokens $n(C_t \setminus G_t)$, the stale subset of those tokens, and the
unneeded tokens remaining after StateTape. \emph{Right:} tokens needed by the next edit $n(G_t)$,
those already held $n(C_t \cap G_t)$, and those held after StateTape
$n(C_t^{+} \cap G_t)$.}
\label{fig:removedretrieved}
\vspace{-3mm}
\end{figure}

\begin{table}[t]
\centering
\scriptsize
\setlength{\tabcolsep}{2.5pt}
\renewcommand{\arraystretch}{0.92}
\begin{tabular}{@{}llrrr@{\hspace{16pt}}llrrr@{}}
\toprule
\textbf{Main agent} & \textbf{Configuration} & \textbf{SWE-EVO} & \textbf{SlopCode} & \textbf{SWE-Pro} &
\textbf{Main agent} & \textbf{Configuration} & \textbf{SWE-EVO} & \textbf{SlopCode} & \textbf{SWE-Pro} \\
\midrule
\multirow{6}{*}{Haiku 4.5} & Control & 25.0 & 4.6 & 26.3 & \multirow{6}{*}{GPT-5.6 luna} & Control & 56.3 & 4.1 & 64.3 \\
 & Obs.\ masking & 27.1 & 5.1 & 27.1 &  & Obs.\ masking & 58.3 & 4.6 & 65.8 \\
 & LLM summary & 27.1 & 6.1 & 26.7 &  & LLM summary & 56.3 & 5.1 & 64.7 \\
 & CORVUS & 25.0 & 5.6 & 26.3 &  & CORVUS & 58.3 & 6.1 & 64.3 \\
 & Text-only mgr & 25.0 & 7.1 & 27.1 &  & Text-only mgr & 58.3 & 5.1 & 66.2 \\
 & \cellcolor{oursC}\textit{StateTape} & \cellcolor{oursC}\textbf{29.2} & \cellcolor{oursC}\textbf{8.1} & \multicolumn{1}{>{\columncolor{oursC}[\tabcolsep][\tabcolsep]}r@{\hspace{16pt}}}{\textbf{27.8}} &  & \cellcolor{oursC}\textit{StateTape} & \cellcolor{oursC}\textbf{60.4} & \cellcolor{oursC}\textbf{6.6} & \multicolumn{1}{>{\columncolor{oursC}[\tabcolsep][0pt]}r@{}}{\textbf{67.7}} \\
\midrule
\multirow{6}{*}{Sonnet 5} & Control & 41.7 & 9.2 & 63.5 & \multirow{6}{*}{GPT-5.6 terra} & Control & 52.1 & 6.6 & 65.8 \\
 & Obs.\ masking & 41.7 & 10.2 & 65.0 &  & Obs.\ masking & 54.2 & 7.1 & 67.7 \\
 & LLM summary & 43.8 & 9.7 & 63.9 &  & LLM summary & 54.2 & 7.7 & 66.2 \\
 & CORVUS & 45.8 & 11.2 & 63.5 &  & CORVUS & 52.1 & 7.1 & 65.8 \\
 & Text-only mgr & 43.8 & 10.7 & 65.4 &  & Text-only mgr & 56.3 & 7.1 & 67.7 \\
 & \cellcolor{oursC}\textit{StateTape} & \cellcolor{oursC}\textbf{47.9} & \cellcolor{oursC}\textbf{12.8} & \multicolumn{1}{>{\columncolor{oursC}[\tabcolsep][\tabcolsep]}r@{\hspace{16pt}}}{\textbf{68.4}} &  & \cellcolor{oursC}\textit{StateTape} & \cellcolor{oursC}\textbf{58.3} & \cellcolor{oursC}\textbf{8.2} & \multicolumn{1}{>{\columncolor{oursC}[\tabcolsep][0pt]}r@{}}{\textbf{68.4}} \\
\midrule
\multirow{6}{*}{Opus 4.8} & Control & 58.3 & 9.7 & 69.2 & \multirow{6}{*}{GPT-5.6 sol} & Control & 58.3 & 8.2 & 64.6 \\
 & Obs.\ masking & 58.3 & 10.7 & 71.1 &  & Obs.\ masking & 60.4 & 8.7 & 69.2 \\
 & LLM summary & 58.3 & 11.2 & 69.5 &  & LLM summary & 58.3 & 9.2 & 67.7 \\
 & CORVUS & 60.4 & 10.7 & 69.2 &  & CORVUS & 58.3 & 9.7 & 67.3 \\
 & Text-only mgr & 60.4 & 12.2 & 68.4 &  & Text-only mgr & 60.4 & 9.7 & 68.1 \\
 & \cellcolor{oursC}\textit{StateTape} & \cellcolor{oursC}\textbf{62.5} & \cellcolor{oursC}\textbf{12.8} & \multicolumn{1}{>{\columncolor{oursC}[\tabcolsep][\tabcolsep]}r@{\hspace{16pt}}}{\textbf{71.4}} &  & \cellcolor{oursC}\textit{StateTape} & \cellcolor{oursC}\textbf{62.5} & \cellcolor{oursC}\textbf{10.2} & \multicolumn{1}{>{\columncolor{oursC}[\tabcolsep][0pt]}r@{}}{\textbf{69.5}} \\
\bottomrule
\end{tabular}
\vspace{-2mm}
\caption{End-to-end resolve rate (\%) on SWE-EVO, SlopCodeBench (SlopCode) and SWE-bench Pro
(SWE-Pro), reported for six main agents from the Claude and GPT-5.6 families. Each main agent is
run under Control, four context-management baselines and StateTape (shaded). Within each
main-agent block, bold denotes the best result on each benchmark.}
\label{tab:main}
\vspace{-3mm}
\end{table}

\vspace{-3mm}

\paragraph{StateTape evicts and retrieves more selectively than other methods.}
On TraceBench (Table~\ref{tab:tbmetrics}), StateTape reaches the highest CE (6.18) and RE (10.47). It clears 74.2\% of unneeded tokens and recovers 68.1\% of
$G_t \setminus C_t$ with 6{,}502 added tokens per decision point, at every edit count
(Figure~\ref{fig:removedretrieved}).

\vspace{-4mm}
\subsection{Ablation Studies}
\label{sec:ablation}

\vspace{-2mm}
Table~\ref{tab:ablation} replaces one component at a time (main agent Sonnet 5): files replace
symbols in the code graph, a text judgment replaces the tape, and default actions the manager.

Removing the code graph causes the largest drop ($-3.4$). CE halves to 3.14, because a file-level
tape cannot tell an edit to one method from an edit that only shifts the definitions below it.
Coverage falls to 44.3\%, because retrieval can no longer follow call edges to the 2{,}721 of
3{,}545 gold regions that are not the edit site.

\begin{table}[t]
\centering
\scriptsize
\setlength{\tabcolsep}{8pt}
\begin{tabular}{@{}lrrr@{}}
\toprule
\textbf{O1: eviction} & Cleared (\%) & Lost / DP & CE $\uparrow$ \\
\midrule
\textsc{keep-all} & 0.0 & 0 & -- \\
\textsc{evict-all} & 100.0 & 4{,}823 & 2.75 \\
recency, $k = 5$ & 96.8 & 4{,}400 & 2.92 \\
recency, $k = 10$ & 91.5 & 4{,}188 & 2.90 \\
recency, $k = 20$ & 80.1 & 3{,}891 & 2.73 \\
recency, $k = 40$ & 57.9 & 3{,}371 & 2.28 \\
age window, 25 & 73.3 & 3{,}733 & 2.61 \\
age window, 50 & 47.1 & 3{,}125 & 2.00 \\
token budget, 16k & 23.6 & 2{,}678 & 1.17 \\
token budget, 32k & 0.4 & 1{,}396 & 0.04 \\
evict non-code & 45.5 & 2{,}430 & 2.49 \\
span-overlap & 22.1 & 626 & 4.69 \\
span-overlap + non-code & 67.6 & 3{,}056 & 2.94 \\
\rowcolor{oursC}
\textit{StateTape} & \textbf{74.2} & \textbf{1{,}595} & \textbf{6.18} \\
\midrule[\heavyrulewidth]
\textbf{O2: retrieval} & Found (\%) & Added / DP & RE $\uparrow$ \\
\midrule
\textsc{retrieve-nothing} & 0.0 & 0 & -- \\
whole next-edit file & 38.4 & 8{,}692 & 4.42 \\
whole files in context & 50.9 & 64{,}592 & 0.79 \\
\rowcolor{oursC}
\textit{StateTape} & \textbf{68.1} & \textbf{6{,}502} & \textbf{10.47} \\
\bottomrule
\end{tabular}
\vspace{-1mm}
\caption{CE and RE on TraceBench. Cleared: unneeded tokens cleared; Lost: needed tokens lost;
Found: missing tokens retrieved; DP: decision point.}
\label{tab:tbmetrics}
\end{table}
Removing the tape costs $-3.1$. Its CE of 2.86 is close to the 2.75 of \textsc{evict-all}, since a
model reading a record's wording cannot tell whether a later write falsified it. Retrieval also
drops to 21.7\%, since it has no drifted symbol to start from.
Removing the manager costs the least ($-2.0$), since the tape's default actions still run. CE stays
at 4.91, but finished records survive and coverage falls to the 31.2\% of \textsc{refresh}.

\begin{table}[t]
\centering
\scriptsize
\setlength{\tabcolsep}{6pt}
\begin{tabular}{@{}lrrrrrrr@{}}
\toprule
 & \multicolumn{4}{c}{\textbf{Resolve rate (\%)}} & \multicolumn{3}{c}{\textbf{TraceBench}} \\
\cmidrule(lr){2-5}\cmidrule(l){6-8}
\textbf{Variant} & SWE-EVO & SlopCode & SWE-Pro & $\Delta$ & CE $\uparrow$ & Missing retrieved (\%) & RE $\uparrow$ \\
\midrule
\rowcolor{oursC}
\textit{StateTape} (full) & \textbf{47.9} & \textbf{12.8} & \textbf{68.4} & --- & \textbf{6.18} & \textbf{68.1} & \textbf{10.47} \\
w/o code graph (file-level tape) & 43.8 & 10.2 & 65.0 & $-3.37$ & 3.14 & 44.3 & 4.86 \\
w/o tape (text-judged staleness) & 43.8 & 10.7 & 65.4 & $-3.07$ & 2.86 & 21.7 & 3.94 \\
w/o manager (tape defaults only) & 45.8 & 11.2 & 66.2 & $-1.97$ & 4.91 & 31.2 & 4.57 \\
\bottomrule
\end{tabular}
\vspace{-2mm}
\caption{Ablation of StateTape's three core components. $\Delta$ is the mean change in resolve rate over the three benchmarks. Bold
denotes the best result.}
\label{tab:ablation}
\vspace{-3mm}
\end{table}

\begin{table}[t]
\centering
\scriptsize
\setlength{\tabcolsep}{2.5pt}
\begin{tabular}{@{}llrrr@{\hspace{14pt}}llrrr@{}}
\toprule
\textbf{Main agent} & \textbf{Benchmark} & \textbf{Agent} & \textbf{Manager} & \shortstack{\textbf{Manager/}\\\textbf{Agent (\%)}} &
\textbf{Main agent} & \textbf{Benchmark} & \textbf{Agent} & \textbf{Manager} & \shortstack{\textbf{Manager/}\\\textbf{Agent (\%)}} \\
\midrule
\multirow{3}{*}{Haiku 4.5}
 & SWE-EVO & 206.9 & 8.4 & 4.1 & \multirow{3}{*}{GPT-5.6 luna} & SWE-EVO & 70.3 & 0.5 & 0.8 \\
 & SlopCodeBench & 526.7 & 19.8 & 3.8 & & SlopCodeBench & 95.2 & 1.9 & 2.0 \\
 & SWE-bench Pro & 1039.3 & 24.0 & 2.3 & & SWE-bench Pro & 206.3 & 1.6 & 0.8 \\
\midrule
\multirow{3}{*}{Sonnet 5}
 & SWE-EVO & 185.8 & 6.1 & 3.3 & \multirow{3}{*}{GPT-5.6 terra} & SWE-EVO & 44.2 & 0.4 & 1.0 \\
 & SlopCodeBench & 575.9 & 4.8 & 0.8 & & SlopCodeBench & 76.5 & 1.8 & 2.3 \\
 & SWE-bench Pro & 730.9 & 22.9 & 3.1 & & SWE-bench Pro & 150.8 & 2.7 & 1.8 \\
\midrule
\multirow{3}{*}{Opus 4.8}
 & SWE-EVO & 77.6 & 2.9 & 3.7 & \multirow{3}{*}{GPT-5.6 sol} & SWE-EVO & 43.4 & 0.5 & 1.1 \\
 & SlopCodeBench & 94.3 & 4.5 & 4.7 & & SlopCodeBench & 73.9 & 0.8 & 1.1 \\
 & SWE-bench Pro & 444.2 & 18.9 & 4.3 & & SWE-bench Pro & 151.0 & 2.8 & 1.8 \\
\bottomrule
\end{tabular}
\vspace{-2mm}
\caption{Token cost of StateTape, in millions of tokens summed over each benchmark's instances
(SlopCodeBench: all checkpoints of the 36 problems). \emph{Agent} counts the main agent's input
tokens, cache reads included, plus output tokens. \emph{Manager} counts the manager's input and
output tokens (Haiku 4.5 for the Claude agents, GPT-5.4 nano for the GPT agents), including calls
made in retried attempts. \emph{Manager/Agent} is the manager's tokens as a percentage of the
main agent's.}
\label{tab:cost}
\vspace{-5mm}
\end{table}
\vspace{-4mm}
\section{Conclusion}
\vspace{-3mm}
Context management is a natural remedy for running coding agents over long horizons, since it bounds what the agent carries across requests. However, a coding agent falsifies its own evidence at every write, and history-based maintenance cannot tell from the text alone which records are stale and which code is needed. In this paper, we present StateTape, a novel and scalable framework that rewrites a coding agent's context at every write rather than by age or size. We model the repository as a symbol-level code graph and build a tape of graph diffs upon it, which turns staleness from an inference about text into an observation of the agent's writes. At each write, the tape nominates the records the write could have falsified, and a manager subagent built on a small LLM settles what the write log alone cannot. We also analyze StateTape theoretically and build TraceBench, a benchmark that labels what an agent holds against what it needs. Extensive experiments with six coding agents on three edit-heavy benchmarks demonstrate that StateTape clears falsified records and retrieves needed code effectively. It thus attains the highest resolve rate in every setting at marginal cost.
\section*{AI use statement}

In this work, we used generative AI tools solely to polish the writing of the manuscript: grammar
and spelling correction, minor rewording for concision, and LaTeX formatting fixes. We did not use
generative AI tools for research ideation, method design, theoretical analysis, implementation of
StateTape or of the baselines, experiment design or execution, data analysis, or literature search
and citation generation; all such contributions are the authors' own. All AI-assisted edits were
made at the sentence level on text the authors had already written, and every suggestion was
reviewed and accepted or rejected by the authors, who verified that no claim, number, or citation
was altered. Large language models otherwise appear in this work only as the systems under study,
i.e., the coding agents and the manager model evaluated in our experiments. We take responsibility
for the final content of this work, including text, claims or artifacts produced with the aid of
generative AI.

\bibliography{iclr2027_conference}
\bibliographystyle{iclr2027_conference}

\appendix

\section{Proofs for Section~\ref{sec:theory}}
\label{app:proofs}

\paragraph{Notation.}
Write $S(C; w) = \{ r \in C : \neg\phi_r(w) \}$ for the records of $C$ that are false of $w$, so
$S_t = S(C_t; w_t)$, and recall the maintained context $C^{+}_t = (C_t \setminus D_t) \cup A_t$ of
Eq.~\ref{eq:update}. A record is \newterm{tracked} if its referent is a set of symbols (a read or
an edit) or a repository state (a test result); search results and assistant text are untracked and
are left to the manager's off-tape eviction. For a set of records $C$, $C^{\mathrm{tr}}$ denotes its
tracked records. We extend
$\mathcal{N}_t$ to contain every state-referent record with a write in $(\tau_r, t]$, the records
that $\textsc{rerun}$ acts on. The \emph{local view} $\lambda_w(v)$ of a symbol $v$ is its body
$b_w(v)$ together with its outgoing edges in $E_w$, or $\bot$ if $v \notin V_w$. We write
$\widetilde{C}_t$ for the context sent to the main agent at step $t$ and
$\Lambda = \sum_{t=1}^{T} n\bigl(S(\widetilde{C}^{\mathrm{tr}}_t; w_t)\bigr)$ for the stale tracked tokens it
is sent over the run.

\begin{assumption}[Symbol locality]
\label{asm:local}
For every record $r$ with a symbol referent, $\rho_r \subseteq V_{w_{\tau_r}}$, and the truth of $r$
depends on the repository only through the local views of its symbols: if
$\lambda_w(v) = \lambda_{w'}(v)$ for all $v \in \rho_r$, then $\phi_r(w) = \phi_r(w')$.
\end{assumption}

Assumption~\ref{asm:local} is the symbol granularity of \S\ref{sec:tape} stated as a condition on
$\phi_r$. It fails when a function's behavior changes through state that belongs to no symbol, such
as a module-level constant.

Theorem~\ref{thm:tape}(b) follows from the set inclusion, valid for every manager output,
\begin{equation}
S\bigl((C^{+}_t)^{\mathrm{tr}}; w_t\bigr) \;\subseteq\; \bigl(N^{\textsc{keep}}_t \cap S_t\bigr) \,\cup\, S(R^{m}_t; w_t),
\label{eq:floor}
\end{equation}
whose last term is empty when no retrieved region covers a symbol of
$\mathcal{X}_{(0,t]} \cup \partial\mathcal{X}_{(0,t]}$.

Throughout, $\mathcal{X}_u$ and $\partial\mathcal{X}_u$ compare $\Phi(u-1)$ with $\Phi(u)$ as
in Eq.~\ref{eq:driftsets}, and a record is true of the state it was observed in,
$\phi_r(w_{\tau_r}) = \mathrm{true}$.

\paragraph{Theorem~\ref{thm:tape}(a).}
Fix a tracked record $r \in C_t$ with $r \notin \mathcal{N}_t$. If $r$ has a state referent, no write
lands in $(\tau_r, t]$, and since observations leave the state fixed, $w_t = w_{\tau_r}$ and
$\phi_r(w_t) = \phi_r(w_{\tau_r}) = \mathrm{true}$. Otherwise $\rho_r$ is a set of symbols and
$\rho_r \cap (\mathcal{X}_{(\tau_r,t]} \cup \partial\mathcal{X}_{(\tau_r,t]}) = \emptyset$. We show by
induction on $u \in (\tau_r, t]$ that every $v \in \rho_r$ satisfies $v \in V_u$ and
$\lambda_{w_u}(v) = \lambda_{w_{u-1}}(v)$. By Assumption~\ref{asm:local}, $v \in V_{\tau_r}$, and by
the induction hypothesis $v \in V_{u-1}$. Since $v \notin \mathcal{X}_u$, $v$ was not removed, so
$v \in V_u$, and its body is unchanged, $b_u(v) = b_{u-1}(v)$. If an outgoing edge $(v, x)$ were
removed from $E_{u-1}$ or added to $E_u$, then $v$ would be an endpoint of a changed edge that lies in
$V_{u-1} \cap V_u$ and outside $\mathcal{X}_u$, so $v \in \partial\mathcal{X}_u$, a contradiction.
Hence $\lambda_{w_u}(v) = \lambda_{w_{u-1}}(v)$. Chaining over $u$ gives
$\lambda_{w_t}(v) = \lambda_{w_{\tau_r}}(v)$ for every $v \in \rho_r$, and
Assumption~\ref{asm:local} gives $\phi_r(w_t) = \phi_r(w_{\tau_r}) = \mathrm{true}$, so
$r \notin S_t$. The argument uses only that the local views at $\tau_r$ and $t$ agree. A rule that
compares each record's outgoing edges at the two ends of its window, as the implementation does, is
therefore sound as well, and the incoming-edge part of $\partial\mathcal{X}$ only adds nominations.

\paragraph{Theorem~\ref{thm:tape}(b).}
Split $C^{+}_t$ into $C_t \setminus D_t$ and $A_t$. Take a tracked $r \in C_t \setminus D_t$. If
$r \notin \mathcal{N}_t$, then $r \notin S_t$ by (a). If $r \in \mathcal{N}_t$, then
$r$ received exactly one action, the manager's or the default. Since $D_t$ contains
$N^{\textsc{drop}}_t \cup N^{\textsc{refresh}}_t \cup N^{\textsc{rerun}}_t$, $r \in N^{\textsc{keep}}_t$.
The off-tape evictions $D^{m}_t$ only remove records. Now take $r \in A_t$. A tombstone states that
the named symbols changed after $\tau_r$, a fact about writes already made, so it holds at $w_t$. A
refreshed body is sliced from $w_t$, so it is true of $w_t$. A $\textsc{rerun}$ stub is an
instruction and makes no claim about the repository. The remaining records of $A_t$ are $R^{m}_t$,
which contribute $S(R^{m}_t; w_t)$. This proves Eq.~\ref{eq:floor}, and the token bound follows by
applying $n(\cdot)$. A retrieved region is a symbol-referent record observed in $w_0$, so
$\tau_r = 0$; if its symbols avoid $\mathcal{X}_{(0,t]} \cup \partial\mathcal{X}_{(0,t]}$, part (a)
applied to it shows that it is true of $w_t$.

\paragraph{Theorem~\ref{thm:cost}(a).}
The drift sets are computed from parse trees and resolved call edges, so no model is called. An
observation leaves $w$ unchanged, so $\mathcal{X}_u = \partial\mathcal{X}_u = \emptyset$, and a record
that enters on an observation step has an empty window. A write changes bytes only in the files it
touches, and whether a symbol exists and what its body is are functions of its own file, so
updating the index of Eq.~\ref{eq:last} requires parsing only the written files. With the index
stored in a hash map, $v \in \mathcal{X}_{(\tau_r,t]}$ holds exactly when $\mathrm{last}_t(v) > \tau_r$,
one lookup per $(r, v)$ pair. For the edge term, the remark at the end of the proof of
Theorem~\ref{thm:tape}(a) lets each record compare the outgoing edges of its symbols at $\tau_r$
with those at $t$, which costs $\mathcal{O}(\deg^{+}_{w_t}(v)) \subseteq \mathcal{O}(d)$ per symbol. Edges are a function of the tree,
so they are re-resolved only after a write. Summing over held records gives the bound.

\paragraph{Theorem~\ref{thm:cost}(b).}
Let a manager call at step $t$ cost at most $\gamma_0 + \beta\,|\mathcal{N}_t|$ tokens. The manager is
invoked only at steps in $\mathcal{T}_N$, so
$K^{\mathrm{mgr}} \le \sum_{t \in \mathcal{T}_N} (\gamma_0 + \beta\,|\mathcal{N}_t|) = |\mathcal{T}_N|\,(\gamma_0 + \beta\,\bar{N})$.
Each request sends its whole context, cache reads included, and compaction is disabled, so
$K^{\mathrm{agt}} \ge \sum_{t=1}^{T} n(\widetilde{C}_t) \ge T\, n_{\min}$. Dividing gives
$K^{\mathrm{mgr}}/K^{\mathrm{agt}} \le \frac{|\mathcal{T}_N|}{T} \cdot \frac{\gamma_0 + \beta\,\bar{N}}{n_{\min}}
\le (\gamma_0 + \beta)\, \frac{|\mathcal{T}_N|\,\bar{N}}{T\, n_{\min}}$, using $|\mathcal{N}_t| \ge 1$
on $\mathcal{T}_N$, which is
the bound in Theorem~\ref{thm:cost}(b).
The live manager meets the hypothesis: its prompt holds one row per nomination (location and at
most five symbol names), previews of at most 40 other records at 80 characters each, and the first
1{,}500 characters of the task, and its reply is capped at 2{,}000 tokens, so $\gamma_0$ does not
grow with the context.

\paragraph{Theorem~\ref{thm:cost}(c).}
Without maintenance no record leaves the context, and by Eq.~\ref{eq:monotone} a stale record stays
stale. Order the writes $s_1 < \dots < s_W$. The at least $\delta$ tokens that write $s_k$ falsifies
are therefore in every request after $s_k$, and there are at least $W - k + 1$ such requests, the
one that returns the result of $s_k$ and one per later write. Hence
$\Lambda \ge \sum_{k=1}^{W} \delta\,(W - k + 1) = \delta\, W(W+1)/2 = \Omega(\delta W^{2})$. With StateTape, staleness arises
only at writes, the context of every request after a write is the maintained $C^{+}_t$, and records
added by observations since then are true of the unchanged state. Theorem~\ref{thm:tape}(b) bounds
each term of $\Lambda$, and summing gives $\Lambda \le \sum_t n(N^{\textsc{keep}}_t \cap S_t)$ when
retrievals cover no drifted symbol. The default policy keeps no nomination and retrieves nothing, so
$\Lambda = 0$.

\section{Method Details}
\label{app:method}

\paragraph{Maintaining the code graph.}
The graph is maintained per write rather than rebuilt per task. A write changes bytes only in the
files it touches, and whether a symbol exists and what its body is are functions of its own file, so
only the written files are re-parsed. Call edges are re-resolved only after a write, and the edge
comparison is restricted to edges incident to a symbol in a written file, so nondeterminism in call
resolution, which lives in files nothing touched, cannot register as change. Appendix~\ref{app:impl}
gives the per-symbol index that answers every query of the tape exactly from these incremental
updates, without materializing the graph at each write.

\paragraph{Drift windows.}
The tape's entries are per-write increments rather than a running comparison against the repository
the run started from, so every change is attributed to the write that made it. For a fixed left
endpoint, the window union $\mathcal{X}_{(s,t]}$ only grows in $t$, which is Eq.~\ref{eq:monotone}
restated at symbol granularity. Symbols in $V_t \setminus V_{t-1}$ are excluded from
Eq.~\ref{eq:driftsets}: a symbol the write created did not exist for any earlier record to have
captured, so it cannot falsify one. Records whose referent is a repository state rather than a set
of symbols, such as test results, are nominated whenever a write falls inside their window.

\paragraph{Maintenance actions.}
Agent transcripts append rather than mutate, so the tape acts by rewriting a nominated record in
place. The tombstone left by \textsc{drop} has constant size and names the referent and the symbols
that changed, so the agent learns that it lost the evidence and why, and can re-read rather than
proceed on a gap it cannot see. \textsc{refresh} slices the current bodies of the drifted symbols
from $w_t$; in the terms of \S\ref{sec:formulation}, the referent appears in both $D_t$ and $A_t$.
A test result observed before later writes cannot be repaired by any current text, only by executing
the test again, which is why state referents take \textsc{rerun}. \textsc{keep} is for nominations
that over-fired. The defaults alone form a complete deterministic policy.

\begin{algorithm}[t]
\caption{StateTape maintenance step at a write (nominate and default)}
\label{alg:tape}
\begin{algorithmic}[1]
\Require context $C_t$; tape state after $a_t \in \mathcal{A}_{\mathrm{wr}}$; repository state $w_t$
\Ensure nominations $\mathcal{N}_t$, each paired with a tape default
\State append to the tape the increment $\mathcal{X}_t, \partial\mathcal{X}_t$ that $a_t$ produced \Comment{Eq.~\ref{eq:driftsets}}
\State $\mathcal{N}_t \gets \emptyset$
\For{each held record $r \in C_t$ with a symbol referent}
  \State $H_r \gets \rho_r \cap (\mathcal{X}_{(\tau_r, t]} \cup \partial\mathcal{X}_{(\tau_r, t]})$ \Comment{Eq.~\ref{eq:nominations}}
  \If{$H_r = \emptyset$} \textbf{continue} \EndIf
  \If{$H_r \cap \mathcal{X}_{(\tau_r, t]} \cap V_t \neq \emptyset$}
    \State $\mathcal{N}_t \gets \mathcal{N}_t \cup \{(r, H_r, \textsc{refresh})\}$
  \Else
    \State $\mathcal{N}_t \gets \mathcal{N}_t \cup \{(r, H_r, \textsc{drop})\}$
  \EndIf
\EndFor
\State \Return $\mathcal{N}_t$
\end{algorithmic}
\end{algorithm}

\paragraph{The per-write procedure.}
Algorithm~\ref{alg:tape} runs after every write. Line~1 appends the write's drift sets to the tape,
the only step that reads the repository. Lines~3--4 intersect each held record's referent with the
drift in its own window, and Line~5 skips records that no drifted symbol touches, so a record is
examined only against the writes that followed it. Lines~7--11 pair the nomination with
\textsc{refresh} when some covered symbol still exists and changed its body, so a current body can
be produced, and with \textsc{drop} when every hit is a removed symbol or a caller whose dispatch
moved. The resulting tuple records the symbols that justify the nomination, which is what the
manager later reads.

\paragraph{The manager's evidence.}
The manager is not handed the repository or the transcript. Its skills (Table~\ref{tab:skills},
Appendix~\ref{app:skills}) expose four kinds of evidence: the nominations, with their reasons, their
defaults and the text a \textsc{refresh} would inject; the drift, meaning which symbols the writes
changed, which callers lie within two hops of them in the code graph, and which held records a given
symbol falsifies; the held context, with each record's text and the step and action that produced
it, from which a record that is finished with is told apart from one still in use; and the
base-state source, to check a region before retrieving it. Every skill is answered by the graph and
the tape without a model. No skill returns the reference patch, and none returns a staleness verdict.

\paragraph{Retrieval labels.}
The manager ranks graph candidates, checks the ones it keeps against source, and labels each
retrieval with the rule that admits it: the edit site, a definition the edit references, a site that
must stay consistent with the change, or a contract the change must satisfy. These are the rules G1
to G4 under which TraceBench labels gold regions (Appendix~\ref{app:tracebench}).

\section{Worked Examples}
\label{app:examples}

\begin{figure}[t]
\centering
\begin{tikzpicture}[
  sym/.style={draw, rounded corners=2pt, minimum height=5mm, minimum width=18mm, inner sep=2pt,
              font=\scriptsize\ttfamily},
  held/.style={sym, fill=gray!20},
  rec/.style={draw, minimum height=5mm, minimum width=18mm, inner sep=2pt, font=\scriptsize\sffamily},
  call/.style={-latex, semithick},
  tag/.style={font=\scriptsize\itshape, inner sep=1pt}]
\node[sym]  (a1) at (0,0)      {init\_app};
\node[held] (a2) at (2.3,0)    {read\_config};
\node[sym]  (a3) at (4.75,0.5) {json\_io.load};
\node[sym]  (a4) at (4.75,-0.5){yaml\_io.load};
\node[rec, fill=gray!20] (a5) at (2.3,-1.0) {test output};
\draw[call] (a1) -- (a2);
\draw[call] (a2) -- (a3);
\node[font=\small] at (2.3,1.25) {(a) Before the write};
\begin{scope}[xshift=7.15cm]
\node[sym]  (b1) at (0,0)      {init\_app};
\node[sym, draw=red!75!black, fill=red!15, very thick] (b2) at (2.3,0) {read\_config};
\node[sym]  (b3) at (4.75,0.5) {json\_io.load};
\node[sym]  (b4) at (4.75,-0.5){yaml\_io.load};
\node[rec, fill=gray!20] (b5) at (2.3,-1.0) {test output};
\draw[call] (b1) -- (b2);
\draw[-latex, gray, dashed] (b2) -- node[red!75!black, font=\large, pos=0.45] {$\times$} (b3);
\draw[call, red!75!black, thick] (b2) -- (b4);
\node[tag, red!75!black, above=0pt of b2, xshift=-6mm] {body unchanged, $\in \partial\mathcal{X}_t$};
\node[font=\small] at (2.3,1.25) {(b) After the write: the tape};
\end{scope}
\begin{scope}[xshift=3.575cm, yshift=-3.1cm]
\node[sym, draw=blue!70!black, fill=blue!10] (c1) at (0,0) {init\_app};
\node[sym, draw=gray, dashed, text=gray] (c2) at (2.3,0) {read\_config};
\node[sym]  (c3) at (4.75,0.5) {json\_io.load};
\node[sym, draw=blue!70!black, fill=blue!10] (c4) at (4.75,-0.5){yaml\_io.load};
\node[rec, draw=gray, text=gray] (c5) at (2.3,-1.0) {test output};
\draw[red!75!black] (c5.south west) -- (c5.north east);
\draw[red!75!black] (c5.north west) -- (c5.south east);
\draw[call] (c1) -- (c2);
\draw[call] (c2) -- (c4);
\node[tag, gray!70!black, above=0pt of c2] {\textsc{drop}: tombstone};
\node[tag, blue!70!black, below=0pt of c1] {retrieved (G3)};
\node[tag, blue!70!black, below=0pt of c4] {retrieved (G2)};
\node[tag, red!75!black, left=2pt of c5] {evicted};
\node[font=\small] at (2.3,1.25) {(c) After the update: the manager};
\end{scope}
\end{tikzpicture}
\vspace{-2mm}
\caption{Examples~2 and~3, an import change. Gray fill marks what the agent holds. (a)~The agent
holds \texttt{read\_config} and the output of an earlier test run. (b)~A write changes the import
behind \texttt{load()}. The body of \texttt{read\_config} is unchanged, but its call edge moves from
\texttt{json\_io.load} to \texttt{yaml\_io.load}, so it enters $\partial\mathcal{X}_t$ and is
nominated. (c)~The manager keeps the default \textsc{drop} and leaves a tombstone, evicts the test
output that no later step uses, and retrieves the new callee (G2) and the caller \texttt{init\_app}
(G3).}
\label{fig:ex-import}
\vspace{-3mm}
\end{figure}
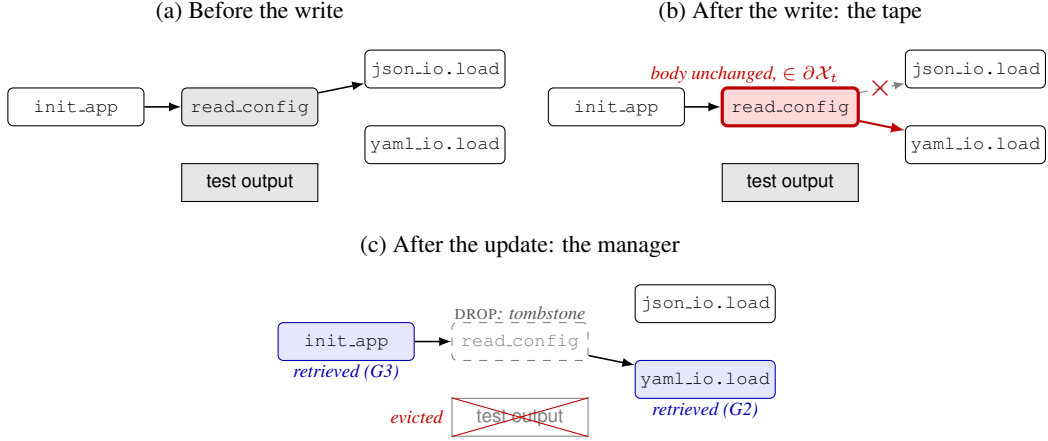

The examples below illustrate the tape (\S\ref{sec:tape}) and the manager (\S\ref{sec:controller}).
They are constructed for exposition rather than taken from a logged run.
Figures~\ref{fig:ex-body} and~\ref{fig:ex-import} draw them in the style of Figure~\ref{fig:example}.

\begin{figure}[t]
\centering
\begin{tikzpicture}[
  sym/.style={draw, rounded corners=2pt, minimum height=5mm, minimum width=18mm, inner sep=2pt,
              font=\scriptsize\ttfamily},
  held/.style={sym, fill=gray!20},
  flow/.style={-latex, thick, gray!60},
  tag/.style={font=\scriptsize\itshape, inner sep=1pt}]
\node[held] (p1) at (0,0) {parse\_config};
\node[tag, below=0pt of p1] {held as record $r$};
\node[font=\small] at (0,1.0) {(a) Read at step $\tau_r$};
\node[sym, draw=red!75!black, fill=red!15, very thick] (p2) at (4.6,0) {parse\_config};
\node[tag, red!75!black, above=0pt of p2] {body edited};
\node[tag, red!75!black, below=0pt of p2] {$\in \mathcal{X}_t$, so $r \in \mathcal{N}_t$};
\node[font=\small] at (4.6,1.0) {(b) Write at step $t$};
\node[held, very thick] (p3) at (9.2,0) {parse\_config};
\node[tag, above=0pt of p3] {\textsc{refresh}};
\node[tag, below=0pt of p3] {$r$ holds the new body};
\node[font=\small] at (9.2,1.0) {(c) After the update};
\draw[flow] ([xshift=3mm]p1.east) -- node[tag, above=1pt] {write $a_t$} ([xshift=-3mm]p2.west);
\draw[flow] ([xshift=3mm]p2.east) -- node[tag, above=1pt] {tape} ([xshift=-3mm]p3.west);
\end{tikzpicture}
\vspace{-2mm}
\caption{Example~1, a body edit. Gray fill marks what the agent holds. (a)~The agent reads
\texttt{parse\_config}, which enters the context as record $r$. (b)~A later write edits its body, so
\texttt{parse\_config} $\in \mathcal{X}_t$ and $r$ is nominated. (c)~Since the symbol still exists,
the tape assigns \textsc{refresh} and $r$ is replaced by the current body.}
\label{fig:ex-body}
\vspace{-3mm}
\end{figure}
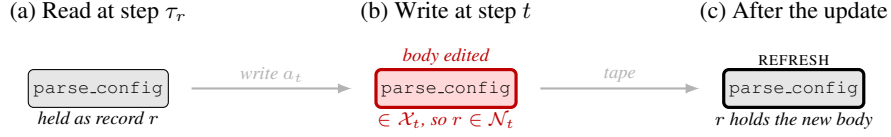

\paragraph{Example 1: a body edit.}
Early in a run the agent reads \texttt{parse\_config(path)}, whose body returns the parsed
dictionary, and the read enters the context as a record $r$ whose referent covers
\texttt{parse\_config}. Several steps later the agent edits that body to validate the dictionary
against a schema and raise on missing keys. The symbol exists in both states and its body differs,
so \texttt{parse\_config} $\in \mathcal{X}_t$ by Eq.~\ref{eq:driftsets}. The write lies inside the
window $(\tau_r, t]$, so $r \in \mathcal{N}_t$ by Eq.~\ref{eq:nominations}. Because the symbol still
exists and its body changed, Algorithm~\ref{alg:tape} assigns \textsc{refresh}, and the
record's text is replaced by the current body, so the agent no longer holds a version that never
raises.

\paragraph{Example 2: caller drift.}
Suppose instead the agent edits a module and changes the line \texttt{from .json\_io import load} to
\texttt{from .yaml\_io import load}. A function \texttt{read\_config} in that module that calls
\texttt{load()} is byte for byte what it was, and a record holding its text still matches the file.
The behavior the record describes has nevertheless changed, because the call now reaches a different
function. The import line belongs to no symbol body, so $\mathcal{X}_t$ does not register the write.
The edge from \texttt{read\_config} moves from \texttt{json\_io.load} to \texttt{yaml\_io.load}, so
the pair lies in $(E_{t-1} \setminus E_t) \cup (E_t \setminus E_{t-1})$ and
$\partial\mathcal{X}_t$ contains \texttt{read\_config}.
The record is nominated, and since its only hit is in $\partial\mathcal{X}_t$, the default is
\textsc{drop}: re-slicing the unchanged body would restore the same text, while the tombstone tells
the agent that the call now resolves elsewhere. Any change that alters dispatch without touching the
caller's text, such as a changed import, a new override in a subclass or a removed definition that a
call now falls through to, reaches the tape through $\partial\mathcal{X}_t$ in the same way.

\paragraph{Example 3: the manager after the import change.}
Continuing Example 2, the manager is invoked because the tape carries a nomination, and it makes the
three decisions of \S\ref{sec:controller}. For the nomination, it accepts the default \textsc{drop} of
\texttt{read\_config}. For eviction, it finds through the held-context skills the output of a test
run from an earlier fix to an unrelated module; no write falsified it, so the tape does not nominate
it, but no later step uses it, and the manager adds it to $D^{m}_t$. For retrieval, it starts from
\texttt{read\_config} and follows call edges: the new callee \texttt{yaml\_io.load} is a definition
the next edit references (G2), and \texttt{init\_app}, which calls \texttt{read\_config} and indexes the
returned dictionary must stay consistent with the new format (G3). It checks both against source and
adds them to $R^{m}_t$. The resulting update is Eq.~\ref{eq:update} with one tombstone, one eviction
and two retrievals.

\section{Implementation Details}
\label{app:impl}

\paragraph{Baselines.}
All four baselines are implemented as modes of the same shim as StateTape.
\textbf{Observation masking} replaces tool outputs older than the last 10 turns with a placeholder.
\textbf{LLM summary} folds the history into a summary once 21 turns have accumulated and keeps the
last 10. It uses the main agent's own model. Both follow the settings of
\citet{lindenbauer2025complexity}. \textbf{CORVUS} \citep{zheng2026corvus} registers every file the
agent reads and re-injects its current contents on each request. The \textbf{text-only manager}
follows the style of AgentDiet \citep{xiao2026reducing}. It uses the same manager model as
StateTape, but it sees only previews of recent reads and searches, and it can only delete them.

\paragraph{Execution environment.}
Each instance runs in its own Docker container with a 90-minute limit (30 minutes per SlopCodeBench
checkpoint) and no turn limit. All runs use a single host with an Apple M5 Pro (18 CPU cores) and
48~GB of memory.

\paragraph{Evaluating a window without keeping the graphs.}
Materializing $\Phi$ at both ends of every write states Eq.~\ref{eq:driftsets} directly, but it
would put a repository-wide static analysis inside the agent's request path, and a window query
would need every graph in the run rather than two. Neither is necessary. Each write is parsed only
for the symbols it touched, and those are folded into a per-symbol last-write index
\begin{equation}
\mathrm{last}_t(v) \;=\; \max\bigl\{\, s \le t \;:\; v \in \mathcal{X}_s \,\bigr\},
\qquad \mathrm{last}_t(v) = -\infty \ \text{ if no write has touched } v ,
\label{eq:last}
\end{equation}
which is all a window query needs, since $v \in \mathcal{X}_{(s,t]}$ exactly when
$\mathrm{last}_t(v) > s$. The index is an exact evaluator of Eq.~\ref{eq:driftsets}, not an
approximation of it, and it refers to no base state, so files the agent creates mid-run are handled
like any other. The edge term is refreshed only over the symbols currently under watch and only
when the tree has moved since the last rebuild, which is likewise exact, because edges are a
function of the tree and the watched seeds and a step that wrote nothing cannot change them.

\section{TraceBench Annotation and Scoring}
\label{app:tracebench}
\paragraph{Reading Table~\ref{tab:tbstats}.}
\S\ref{sec:gap} discusses what the two regions contain. For scoring, note that the left region is
73.4\% of held tokens at a pre-edit decision, so O1 has a high base rate and $F_1$ alone rewards
indiscriminate deletion; precision at fixed retained-useful mass is the informative view.

\begin{table}[t]
\centering
\small
\setlength{\tabcolsep}{4pt}
\begin{tabular}{@{}lr@{\hspace{2em}}lr@{}}
\toprule
\multicolumn{2}{@{}l@{\hspace{2em}}}{\textbf{Trajectories and decision points}} & \multicolumn{2}{@{}l@{}}{\textbf{Left region} $C_t \setminus G_t$ \textbf{(O1)}} \\
\cmidrule(r){1-2}\cmidrule(l){3-4}
instances / repositories & 48 / 7 & unneeded tokens & 5{,}349{,}888 \\
steps / trajectory (median / max) & 71 / 356 & share of held tokens & 73.4\% \\
gold hunks / instance (median / max) & 21 / 379 & \quad stale & 72.3\% \\
decision points (all pre-edit) & 403 & \quad consumed & 10.4\% \\
instances with $\ge 1$ point & 43 & \quad distractor & 17.3\% \\
depth in steps (median / max) & 75 / 278 & tokens in records touching $G_t$ & 1{,}943{,}649 \\
\midrule
\multicolumn{2}{@{}l@{\hspace{2em}}}{\textbf{Held context} $C_t$} & \multicolumn{2}{@{}l@{}}{\textbf{Right region} $G_t \setminus C_t$ \textbf{(O2)}} \\
\cmidrule(r){1-2}\cmidrule(l){3-4}
context records scored & 36{,}873 & gold tokens & 697{,}727 \\
held tokens & 7{,}293{,}537 & share of gold not held & 58.5\% \\
token encoding & \texttt{cl100k} & coverage $\mathrm{Rec}(C_t)$ & 0.415 \\
judge & \texttt{gpt-5.4}, seed 0 & gold regions G3 / G4 & 1{,}338 / 595 \\
 & & oracle tokens / point & 1{,}012 \\
\bottomrule
\end{tabular}
\caption{TraceBench at a glance. Every quantity is a token count under one pinned encoding. The
two right-hand blocks are the two regions of Figure~\ref{fig:partition} and the two surrogates of
Eq.~\ref{eq:surrogates}. The left region is scored per record, so a held record that overlaps
$G_t$ counts in full and each repeated read counts again. The right region is scored over
deduplicated gold spans. The two intersections $C_t \cap G_t$ therefore differ in size,
1{,}943{,}649 tokens on the left against about 289{,}600 on the right.}
\label{tab:tbstats}
\end{table}

\paragraph{Source trajectories and decision points.}
Each raw event stream is normalized into typed steps by one parser per agent wire format; nothing
downstream of the parser is agent-specific. Sub-agent activity is excluded from $C_t$ on purpose: only a sub-agent's returned summary enters the
main thread, so a file it rewrote and did not surface is genuinely absent from the main agent's
context and is a real source of both regions. Typed steps are
$\{\textsc{read}, \textsc{edit}, \textsc{run}, \textsc{search}, \textsc{say}\}$, carrying either a
$(\mathit{file}, \mathit{span})$ or a command. A decision point $t$ means the context is
$s_1 \ldots s_t$ and the label describes the decision faced at $s_{t+1}$. Three anchor families are
implemented: \textbf{P}, pre-edit, at $t = j-1$ for each edit \emph{burst} beginning at $s_j$, a
maximal run of consecutive edits, so that a ten-line multi-edit to one function is one decision and
not ten; \textbf{T}, post-test, at each genuine test run; and \textbf{F}, final, at $t = n$. The
shipped label set is P only, because every pre-edit decision point has an imminent edit to anchor
$G_t$ on and T and F carry none. Of 551 pre-edit candidates, 148 are skipped where the imminent burst
is purely shell or diagnostic and has no localized edit in a gold file to anchor on. Trajectory
lengths reach 356 steps and reference patches 379 hunks; decision-point depth reaches 278 steps.

\paragraph{Gold regions.}
Each gold region is admitted under one of four rules, cited on the label: \textbf{G1}, the edit site,
the enclosing definition of the imminent hunk in its pre-edit state; \textbf{G2}, a referenced
definition the hunk calls, imports or subclasses; \textbf{G3}, the blast radius, an existing site
that must stay consistent with the change; and \textbf{G4}, the contract, the interface, base class,
registry or schema the change must satisfy. These are the rules the retrieval proposals of
\S\ref{sec:controller} are named for. Four bounds keep the set tight: co-location in a touched file
without a traceable dependency is not gold, mechanical boilerplate is not gold, test code is never
gold since it encodes the answer, and a region already held is the intersection of
Figure~\ref{fig:partition} rather than the right region.

\paragraph{The held context $C_t$.}
$C_t$ is materialized as an inventory of records, each
$\{\mathit{id}, \mathit{kind}, \mathit{file}, \mathit{span}, \mathit{born\_step}, \mathit{text},
n\}$, which is the record $(c_r, \rho_r, \tau_r)$ of \S\ref{sec:formulation} made concrete:
file-region items, from reads and edits, carry a referent, while run outputs, search results and
assistant text do not. All text is tokenized with a pinned \texttt{cl100k\_base} encoding, and
every quantity reported is a token count.

\paragraph{Outstanding hunks.}
A deterministic pre-pass parses the reference patch into source hunks, drops the test patch, and
marks a hunk outstanding at $t$ if no agent edit at or before $t$ overlaps its base span. The judge
then selects, among outstanding hunks, the one the imminent burst is positioned to make, and the
G1--G4 sweep is anchored on that selection rather than on the patch as a whole.

\paragraph{Annotation protocol.}
The needed pass runs a per-file sweep over the repository at the base commit, reading files until
every outstanding gold file is covered and emitting regions with a rule and a justification; the
orchestration around it is deterministic rather than model-chosen, so coverage of the gold files is
guaranteed rather than hoped for. The stale pass is trajectory-framed and works from the item
record inventory with deterministic supersession flags pre-attached. Reproducibility is
pinned throughout: endpoint, deployment, model version, prompt version and seed are echoed in the
manifest, the request cache is content-addressed and authoritative so replay needs no key, and a
prompt change re-namespaces the cache. Labels carry the transcript hash and are byte-reproducible.

\paragraph{Scoring.}
Predictions and labels are expanded to token sets, gold regions from the base-commit file slice and
held records from their observed text, and overlap is resolved at the
$(\mathit{file}, \text{token index})$ level, so a partially read gold region contributes only its
unread tokens. The headline numbers are the two ratios of \S\ref{sec:metrics}, $\mathrm{CE}$
(Eq.~\ref{eq:ce}) against the left region and $\mathrm{RE}$ (Eq.~\ref{eq:re}) against the right,
each micro-averaged over the decision points of a split and each paired with its volume; we also
report the underlying token-weighted precision, recall and $F_1$ for O1 and raw tokens added per
decision for O2, since those are the quantities the ratios are built from.
Reference points are reported with every result: \textsc{keep-all}, \textsc{evict-all}, recency and
age windows, and a span-overlap supersession detector for O1; \textsc{retrieve-nothing},
whole-file heuristics and the oracle for O2.

\paragraph{Provenance and limitations.}
The labels are silver: deterministic rules plus an oracle model, with a human review queue built
and a verification server in place but not yet filled, so scoring a detector measures agreement
with the judge rather than with a person. The labeled set is one agent on one suite, and although
the pipeline is harness-agnostic downstream of the parser, we do not claim the base rates transfer
to a different scaffold. $G_t$ is defined relative to the reference patch, so a correct solution
the reference did not take is scored as not needed; this bounds O2's ceiling rather than distorting
the ranking of policies against each other.

\section{Benchmark Comparison}
\label{app:benchcmp}

Standard evaluation scores a run once, when it ends, while the gap between $C_t$ and $G_t$ moves at
every write. A pass rate can say that a maintenance policy helped without saying which tokens it
should have dropped at step 40, or which region of the repository it should have pulled in before
the edit at step 41. Three families of benchmark bear on the gap. SWE-bench \citep{jimenez2024swebench} and its descendants
\citep{swebenchpro2025,sweevo2025,slopcodebench2025,swemarathon2025} pair a repository issue with
held-out tests and score a run by whether its terminal patch passes. Long-context and memory
benchmarks \citep{maharana2024locomo,locobenchagent2025} measure whether a model can find and use
information placed inside a large given window. Localization benchmarks
\citep{contextbench2026,corebench2026,chen2025locagent} annotate the code an agent should have
found for a task, which is an instantiation of $G_t$ and therefore of the right region.

None of them supervises how an agent maintains its context over a long horizon. First,
outcome-only supervision is too coarse: an instance yields one bit, and that bit is produced by the
agent's model, so a correct eviction the model then ignores and an incorrect one it happens to
survive are indistinguishable at the metric, and the effect of a maintenance policy reaches the
score only through a capability the policy does not control. Our own runs (\S\ref{sec:experiments})
show how loose the coupling is: StateTape rewrites the context at nearly every write, and the
resolve rate moves by a few points, but that one bit per instance never says which of those
rewrites was the right one. Second, the evidence in
long-context benchmarks is static with respect to the model: the distractors were placed by the
benchmark, not caused by the model, and nothing the model does falsifies what it was shown, whereas
a coding agent's own writes falsify what it read and are the only source of its staleness
(\S\ref{sec:formulation}). Third, localization labels are attached to a task rather than to a step,
so a policy that acts at every write has no per-write target and the gold context cannot move as the
repository does; and none of these datasets annotates $C_t \setminus G_t$, so the left region, the
one that carries the token cost, is unmeasured. Their selection criteria also run the other way:
Loc-Bench excludes issues whose patch touches more than five files or ten functions, which is the
right filter for one-shot localization and removes exactly the write-dense instances in which stale
evidence accumulates. Table~\ref{tab:benchcmp} summarizes the comparison.

\begin{table}[t]
\centering
\scriptsize
\setlength{\tabcolsep}{5pt}
\begin{tabular}{@{}lccccc@{}}
\toprule
 & & & \textbf{self-} & \multicolumn{2}{c}{\textbf{annotated regions}} \\
\cmidrule(l){5-6}
\textbf{Family} & \textbf{scored unit} & \textbf{label site} & \textbf{falsifying}
& $C_t \setminus G_t$ & $G_t \setminus C_t$ \\
\midrule
Issue resolution        & terminal patch & task           & yes & no  & no  \\
Long context / memory   & answer         & task           & no  & no  & no  \\
Retrieval / localization& retrieved set  & task           & no  & no  & yes \\
\midrule
\textbf{TraceBench}     & context state  & decision point & yes & yes & yes \\
\bottomrule
\end{tabular}
\caption{Where the evidence lifecycle falls between benchmark families, represented by SWE-bench
and SWE-bench Pro \citep{jimenez2024swebench,swebenchpro2025}, LoCoMo and LoCoBench-Agent
\citep{maharana2024locomo,locobenchagent2025}, and ContextBench and LocAgent
\citep{contextbench2026,chen2025locagent}. \emph{Self-falsifying} asks whether the agent's own
actions can make its evidence wrong, the condition that makes staleness a long-horizon property
rather than a retrieval failure (Eq.~\ref{eq:monotone}). The last two columns are the left and
right regions of Figure~\ref{fig:partition}, what a maintenance policy should evict and what it
should retrieve.}
\label{tab:benchcmp}
\end{table}

\section{Manager Skills}
\label{app:skills}

Table~\ref{tab:skills} lists the skills of \S\ref{sec:controller}: nine read-only
queries answered by the code graph and the tape, and one skill that commits the decision.

\begin{table}[t]
\caption{The manager's skills. Nine read-only queries against the tape, and one that commits the
decision. The manager is given a budget of $25$ calls per invocation and a graph depth of $2$.}
\label{tab:skills}
\centering\small
\begin{tabular}{@{}llp{0.42\linewidth}@{}}
\toprule
\textbf{Group} & \textbf{Skill} & \textbf{Returns} \\
\midrule
\multirow{2}{*}{Nominations}
 & \texttt{list\_candidates} & $\mathcal{N}_t$ with each reason \\
 & \texttt{candidate\_fresh}$(r)$ & $\mathrm{fresh}(r)$, the text a $\textsc{refresh}$ would inject \\
\midrule
\multirow{3}{*}{Drift}
 & \texttt{symbols\_touched} & the symbols the run's writes changed \\
 & \texttt{dependents}$(v,h)$ & callers of $v$  \\
 & \texttt{evidence\_for}$(v)$ & $\{\, r \in C_t : v \in \rho_r \,\}$, the records $v$ falsifies \\
\midrule
\multirow{3}{*}{Context}
 & \texttt{list\_context} & $C_t$ with kind, referent, birth step and token cost \\
 & \texttt{read\_context}$(r)$ & the untruncated text of one held record \\
 & \texttt{provenance}$(r)$ & $\tau_r$ and the action that produced $r$ \\
\midrule
Repository & \texttt{read\_file}$(p,s,e)$ & verify a region before retrieving it \\
\midrule
Decision & \texttt{submit} & $(D^{m}_t,\; R^{m}_t)$ \\
\bottomrule
\end{tabular}
\end{table}

\end{document}